\documentclass[10pt,twocolumn,letterpaper]{article}

\usepackage{cvpr}
\usepackage{times}
\usepackage{epsfig}
\usepackage{graphicx}
\usepackage{amsmath}
\usepackage{amssymb}

\usepackage{multirow}

\usepackage[pagebackref=true,breaklinks=true,letterpaper=true,colorlinks,bookmarks=false]{hyperref}
\usepackage{comment}

\cvprfinalcopy 

\def\cvprPaperID{4365} 

\ifcvprfinal\fi
\begin{document}

\title{Interactively Transferring CNN Patterns for Part Localization}

\author{Quanshi Zhang, Ruiming Cao, Shengming Zhang, Mark Edmonds, Ying Nian Wu, and Song-Chun Zhu\\
University of California, Los Angeles}

\maketitle

\begin{abstract}
This paper explores an interactive method to diagnose knowledge representations of a CNN, in order to use CNN knowledge to model object parts. Unlike traditional end-to-end learning of CNNs that require numerous training samples, given a CNN pre-trained for object classification, we mine object part patterns from the CNN in the scenario of one/multi-shot learning. More specifically, our method uses very few (\emph{e.g.} three) object images to summarize knowledge in conv-layers into a dictionary of latent activation patterns. For each object part, our method visualizes the latent patterns and asks users to manually assemble latent patterns related to the target part, so as to construct the object-part model. As a general solution, our interactive method was broadly applied to different types of neural patterns in experiments. With the guidance of human interactions, our method exhibited superior performance of part localization.\footnote[1]{Quanshi Zhang is the corresponding author. \textit{zhangqs@g.ucla.edu}}
\end{abstract}

\section{Introduction}

Convolutional neural networks (CNNs)~\cite{CNN,CNNImageNet} have shown promise for classification tasks in computer vision. However, learning a model using small data, \emph{e.g.} annotations on 1--3 examples, is still a great challenge for state-of-the-art algorithms. Without sufficient annotations, the conventional end-to-end learning usually has no mechanism to ensure that the CNN actually learns correct knowledge, rather than over-fit to noisy signals.

Therefore, instead of letting the CNN ``guess'' new knowledge from small training data, we aim to mine certain patterns from a well pre-trained CNN to represent semantic parts of the object for part localization. We notice that when a CNN is well pre-trained/finetuned using a large number of object-box annotations for object classification, the CNN has encoded massive implicit patterns for local object shapes in its conv-layers. Each pattern may represent a local shape of an object.

\begin{figure}[t]
\centering
\includegraphics[width=\linewidth]{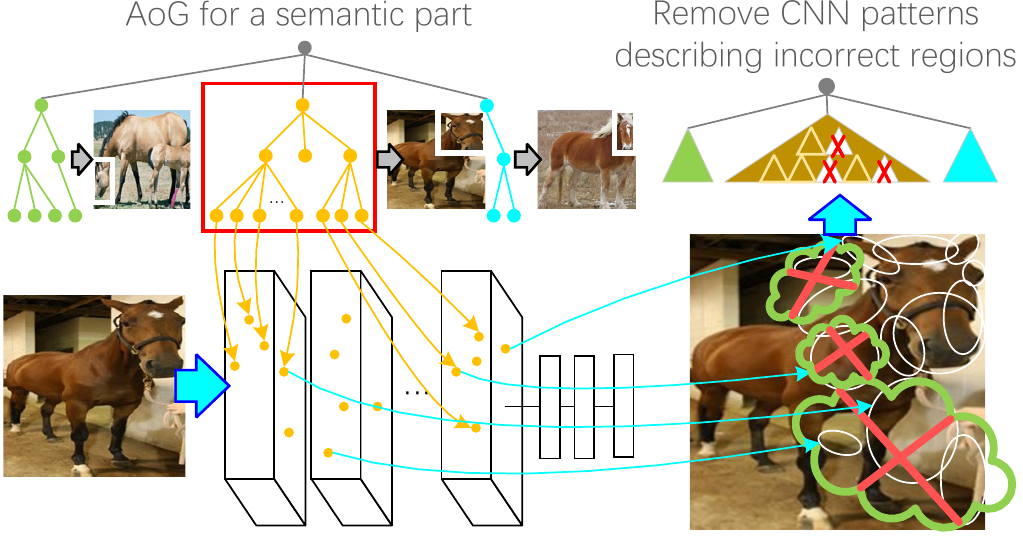}
\caption{Given a pre-trained CNN and very few (\emph{e.g.} \textbf{1--3}) part annotations, we retrieve certain patterns from conv-layers of the CNN to represent the target part. We use an AOG to encode the semantic hierarchy of the retrieved patterns. Each node in the AOG corresponds to a latent pattern in the CNN. We visualize patterns in the AOG and requires people to select patterns related to the target part and remove unrelated patterns, in order to refine the AOG model.}
\label{fig:top}
\end{figure}

In this paper, we propose to incorporate human interactions into the mining of part patterns, in order to ensure we use correct patterns to represent the target semantic part. As shown in Fig.~\ref{fig:top}, we are given a large number of images with object-box annotations to train a CNN for category classification, but very few (\emph{e.g.} 1--3) objects of a category have annotations of a semantic part. We mine hundreds of latent patterns from the pre-trained CNN as candidates that potentially represent the target part. Then, we visualize the mined patterns and ask human users to manually select semantically correct latent patterns to build up the model for part localization, just like playing LEGO blocks.

More specifically, during the process of pattern mining, each latent pattern is expected 1) to frequently appear on objects of the category, 2) to be strongly activated by a compositional (or contextual) shape \emph{w.r.t} the target part, and 3) to keep good spatial relationship with other latent patterns. Because all latent patterns are well pre-trained using massive data, these patterns represent common shapes of a category, instead of being over-fitted to a few part annotations.

For human interactions, we require human users to remove patterns corresponding to background noises and specificity of certain samples. Because we allow people to directly point out model flaws, this interactive-learning strategy is more effective than end-to-end ``guessing'' true knowledge of the object part from small training data.

\textbf{And-Or graph representation:}{\verb| |} Before conducting human interactions, we need to represent latent patterns at the semantic level for pattern manipulation, rather than at the level of CNN neural units. Note that the same object part (\emph{e.g.} the head) in different images may appear in different image positions, thereby activating neurons in different feature map positions. Thus, given a feature map of a filter in a conv-layer, we need to first remove noisy activations from the feature map, and then summarize the rest neural activations on numerous neural units into a few latent patterns for human interactions. When we infer a latent pattern in different images, the pattern may appear in different feature map positions due to object deformation. Moreover, a filter's feature map may be activated by multiple object parts (\emph{e.g.} being activated by both the head and the leg), we need to disentangle latent patterns of different object parts from the same feature map.

\begin{figure*}[t]
\centering
\includegraphics[width=\linewidth]{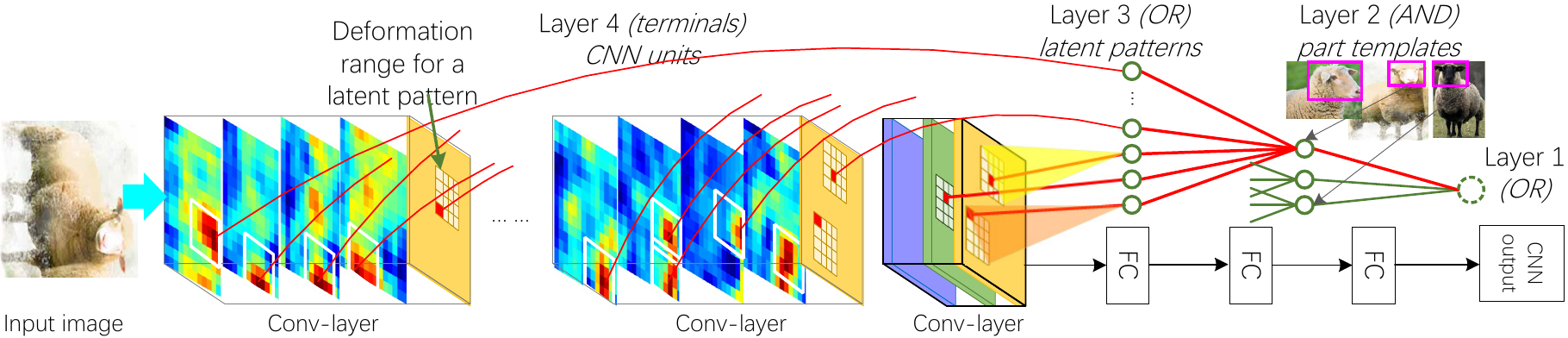}
\caption{Semantic And-Or graph grown on the pre-trained CNN. Red lines in the AOG indicate the parse tree, which associates certain CNN units with certain image regions.}
\label{fig:AOG}
\end{figure*}

Therefore, we use an And-Or graph (AOG) to clearly represent the semantic hierarchy of the patterns that are mined from conv-layers. As shown in Fig.~\ref{fig:AOG}, we build a four-layer AOG to represent the semantic hierarchy ranging from \textit{semantic part}, \textit{part templates}, \textit{latent patterns}, to \textit{CNN units}. We use AND nodes in the AOG to encode compositional regions of a part, and use OR nodes to encode a list of alternative appearances/deformations for a local region.


Based on the AOG, we localize patterns on CNN feature maps and visualize these patterns. Then, users can remove irrelevant patterns by pruning certain AOG nodes.

\textbf{Pattern visualization and interactions:}{\verb| |} We apply the up-convolutional neural network (up-conv-net) in \cite{FeaVisual} to visualize latent patterns in the AOG. We train different up-conv-nets to visualize latent patterns corresponding to different conv-layers of the CNN. According to visualization results, latent patterns in low conv-layers usually describe object details, and those in high conv-layers mainly correspond to large-scale parts or contexts. Therefore, we use low-layer patterns to represent details within the target part, and require users to remove low-layer patterns outside the part. We select patterns in high conv-layers to represent the contextual information of the target part.

\textbf{Method generality:} Our method is a general solution to interactively learning object parts based on pre-trained CNNs. First, the AOG representation of objects~\cite{MiningAOG} can be compatible with different features. Second, our method has been tested on AOGs with two different types of neural patterns~\cite{CNNAoG,explanatoryGraph_arXiv}.

\textbf{Contributions:} Instead of end-to-end learning new information from training data, this study explores the probability of directly selecting certain patterns from a pre-trained CNN to build a model for part localization. Our method mines middle-level latent patterns from the CNN and incorporates human interactions to manually select correct patterns. Our method exhibited superior performance in experiments, which demonstrates the effectiveness of human interactions in weakly-supervised learning.

\section{Related work}

\textbf{CNN visualization, semanticization, and interactions:}{\verb| |} In recent years, many methods have been developed to explain the semantics hidden in the CNN. Studies of \cite{CNNVisualization_1,CNNVisualization_2,CNNVisualization_3} passively visualized content of some given CNN units. \cite{CNNVisualization_5} analyzed statistics of CNN features.

Unlike passive CNN visualization, we hope to actively semanticize CNNs by discovering patterns related to the target part, which is more challenging. Given CNN feature maps, Zhou~\emph{et al.}~\cite{CNNSemanticDeep,CNNSemanticDeep2} discovered latent ``scene'' semantics. Simon~\emph{et al.} discovered objects~\cite{ObjectDiscoveryCNN_2} from CNN activations in an unsupervised manner, and learned part concepts in a supervised fashion~\cite{CNNSemanticPart}. \cite{CNNAoG} mined CNN patterns for a part concept and transformed the pattern knowledge into an AOG model.

However, without sufficient supervision, previous studies cannot ensure the extracted CNN patterns to have correct part semantics. Thus, we visualize candidate patterns and require users to manually select correct ones, so as to create a better ``white-box'' explanation of the target part.


\textbf{And-Or representation:}{\verb| |} In many studies, people used AOGs to represent the hierarchical semantic hierarchy of objects or scenes~\cite{AllenAOG,MiningAOG}. We use the AOG to associate the latent patterns with part semantics, which eases the visualization of CNN patterns and enables semantic-level interactions on CNN patterns.

\textbf{Unsupervised learning of objects and parts:}{\verb| |} Unsupervised object discovery~\cite{ObjectDiscoveryCNN_2} was formulated as a problem of mining common foreground patterns from images, and many sophisticated methods were developed for this problem. Whereas, given a pre-trained object-level model, un-/weakly-supervised learning of part representations is a different problem. 



\section{Algorithm}

In this section, we propose a general method to interactively learn object-part models. In fact, there are a number of techniques to mine neural patterns from CNNs to represent middle-level object shapes. Each of these patterns can be organized using a widely used AOG model. A clear AOG representation allows human users to refine the model by interactively modifying the AOG structure. The basic idea is to visualize latent patterns in the AOG and to let human users identify and remove latent patterns that are not closely related to the target part.

\subsection{Preliminaries: AOG representation}

The AOG has been a typical object representation for years~\cite{MiningAOG,AllenAOG}. Here, we briefly introduce the AOG structure, which has been widely used to represent neural patterns of CNNs in studies of \cite{CNNAoG} and \cite{explanatoryGraph_arXiv}.

The AOG organizes latent patterns hidden in the CNN to explain the semantic hierarchy of an object part. We use the AOG to parse object parts from images. As shown in Fig.~\ref{fig:AOG}, we use a four-layer AOG hierarchy ranging from \textit{semantic part} (OR node), \textit{part templates} (AND nodes), \textit{latent patterns} (OR nodes), to \textit{CNN units} (terminals). In the AOG, an OR node encodes a list of alternative candidates as children, while an AND node uses each of its children to describe a certain compositional shape (or a contextual area) of the father node.

Given an object image $I$\textcolor{red}{\footnote[2]{Considering the CNN's superior performance in object detection, as in \cite{SemanticPart}, object detection and part localization are considered two separate processes for evaluation. Thus, we crop $I$ to only contain the object and resize $I$ for CNN inputs to simplify the scenario of learning for part localization.}}, we use the AOG for part parsing. \emph{I.e.} we first use the CNN to compute $I$'s feature maps of its conv-layers, and then determine a parse tree within the AOG to explain neural activations on the feature maps and simultaneously localize the target part.

As red lines in Fig.~\ref{fig:AOG}, during the parsing procedure, we 1) select a certain part template (AND node in the 2nd Layer) to explain the target part (root OR node in the AOG), 2) parse an image region for the part template (\emph{i.e.} part localization), and 3) for each latent pattern (OR node in the 3rd Layer) under the part template, determine a CNN unit (terminal node) within a deformation range to localize the local shape of this latent pattern. To be precise, we achieve the part parsing in a bottom-up manner. \emph{I.e.} in the beginning, we compute an inference score for each terminal node (CNN unit), and then propagate these scores up to nodes of latent patterns and part templates following certain And-Or rules for part parsing.

The top node of ``semantic part'' (OR node) encodes a number of alternative part templates as children, each denoted by $U\in\Omega^{\textrm{temp}}$. Each part template naturally corresponds to a type of part appearance observed from a certain perspective. Given parsing results of all part templates, the top node selects the part template with the highest inference score as the true parsing configuration.
\begin{equation}
{\bf S}_{I}=\max_{U\in\Omega^{\textrm{temp}}}\max_{\Lambda_{U}}S_{I}(U|\Lambda_{U})
\label{eqn:OR}
\end{equation}
where $S_{I}$ denotes the overall inference score on image $I$, and $\Lambda_{U}$ represents the image region parsed for part template $U$. $S_{I}(U|\Lambda_{U})$ measures the inference score when we parse an image region $\Lambda_{U}$ for the part template $U$.

Then, each part template $U$ (AND node) uses children latent patterns to represent its local compositional shapes or contextual area. Thus, we can formulate $U$'s inference score as the sum of its children's inference scores, \emph{i.e.} $S_{I}(U|\Lambda_{U})=\sum_{V\in\Omega^{U}}S_{I}(V|\Lambda_{V})$, where $\Omega^{U}$ is the children set of $U$, and $S_{I}(V|\Lambda_{V})$ denotes the inference score of $V$.

Finally, each latent pattern $V$ (OR node) naturally corresponds to a square deformation range $R_{V}$ within a certain conv-slice/channel $D_{V}$ of the CNN. All CNN units within $R_{V}$\textcolor{red}{\footnote[3]{We set a constant deformation range $R_{V}$ for each latent pattern, which covers $1/3$-by-$1/3$ of the feature map of the conv-slice. Deformation ranges of different patterns in the same conv-slice may overlap. The central position of $R_{V}$, ${\bf p}(R_{V})$, is a parameter to estimate.}} are regarded as deformation candidates of $V$. Just following the OR-node logic in Eq.~(\ref{eqn:OR}), $V$ also selects the best child unit $\hat{T}$ as the true parsing configuration, $S_{I}(V|\Lambda_{V})=\max_{T\in{R_{V}}}S_{I}(T|\Lambda_{T})$, $\Lambda_{V}=\Lambda_{\hat{T}}$. The selected child $\hat{T}$ propagates both its score and parsed region to parent $V$. The image region $\Lambda_{T}$ for each CNN unit $T$ is fixed, \emph{i.e.} we simply propagate the receptive field of $T$ to the image plane and obtain $\Lambda_{T}$. Therefore, we can re-write the above And-Or logics as a DPM-like (deformable part model) model:
\begin{equation}
\!S_{I}(U|\Lambda_{U})\!=\!\sum_{V\in\Omega^{U}}\max_{\Lambda_{V}}\Big\{S_{I}^{\textrm{loc}}(V|\Lambda_{V})\!+\!\lambda^{\textrm{geo}}S_{I}^{\textrm{geo}}(\Lambda_{V}|\Lambda_{U})\Big\}\!\!
\label{eqn:AND}
\end{equation}
where the part template score comprises latent pattern scores, and each latent pattern $V$ selects its best CNN unit as the parsing configuration (\emph{i.e.} computing the image region corresponding the CNN unit). The unary term $S_{I}^{\textrm{loc}}(V|\Lambda_{V})$\textcolor{red}{\footnote[4]{$S_{I}^{\textrm{loc}}(V|\Lambda_{V})\!=\!S_{I}(\Lambda_{V})\!-\!\lambda^{\textrm{def}}\Vert{\bf p}(\Lambda_{V})\!-\!{\bf p}(R_{V})\Vert^2$ measures both the neural response $S_{I}(\Lambda_{V})$ of the CNN unit corresponding to the position of $\Lambda_{V}$ and the score for local deformation \emph{w.r.t.} $V$'s ideal position ${\bf p}(R_{V})$, where ${\bf p}(\cdot)$ returns the center position. $S_{I}^{\textrm{geo}}(\Lambda_{V}|\Lambda_{U})\!=\!-\lambda^{\textrm{geo}}\Vert{\bf p}(\Lambda_{V})+\Delta_{VU}-{\bf p}(\Lambda_{U})\Vert^2$, where $\Delta_{VU}$ denotes the displacement from $V$ to $U$. $\Delta_{VU}$ is computed based on part annotations. $\lambda^{\textrm{def}}\!=\!1/3$, $\lambda^{\textrm{geo}}\!=\!5.0$.}} measures the local inference quality, and the pairwise term $S_{I}^{\textrm{geo}}(\Lambda_{V}|\Lambda_{U})$\textcolor{red}{\footnotemark[4]} encodes spatial relationship between part template $U$ and latent pattern $V$.

\begin{figure*}[t]
\centering
\includegraphics[width=\linewidth]{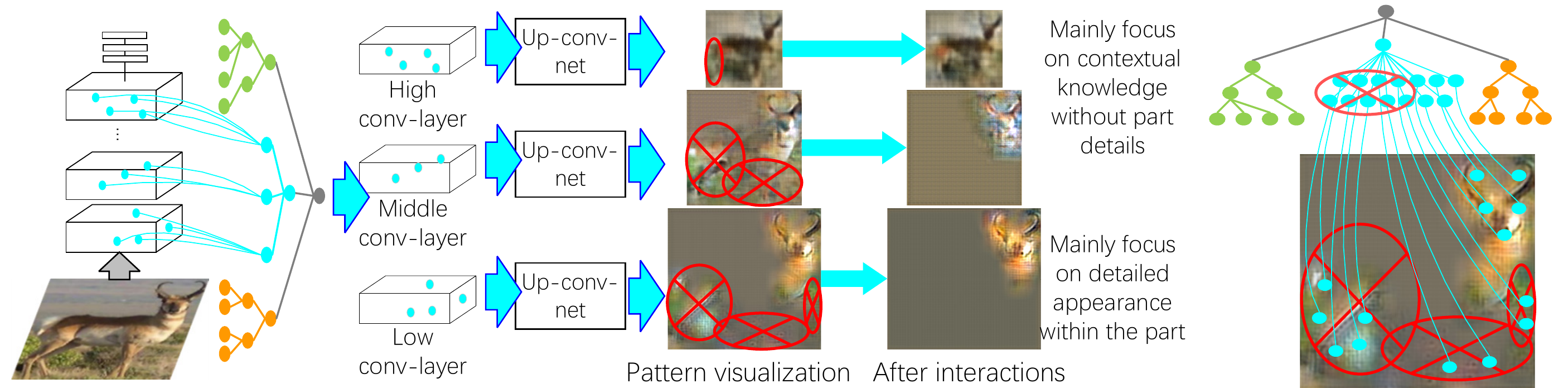}
\caption{Human interactions. Users remove latent-pattern nodes, which are not related to the target part, from the AOG.}
\label{fig:visual}
\end{figure*}

\subsection{Latent patterns}

In order to demonstrate the generality of the proposed method, we use two different types of latent patterns mined from CNNs to construct different AOGs. Our method allows users interactively select different latent patterns for learning. The first type of patterns are proposed in \cite{CNNAoG} as local middle-level features of objects, namely \textit{middle-level patterns}. The method of \cite{explanatoryGraph_arXiv} is proposed to learn an explanatory graph to explain the knowledge hierarchy inside a pre-trained CNN. We use nodes in the explanatory graph as the second type of patterns, namely \textit{explanatory patterns}. In experiments, we compared part-localization performance of AOGs based on different patterns.

The experimental settings for the mining of the above two patterns are the same, which can be summarized as follows. The input is a set of cropped object images of a category, denoted by ${\bf I}$, where only a few objects ${\bf I}^{\textrm{ant}}\!=\!\{I_{i}|i\!=\!1,2,\ldots,M\}\subset{\bf I}$ have ground-truth annotations of the target part. Each part annotation on image $I\!\in\!{\bf I}^{\textrm{ant}}$ includes both the ground-truth part template $U^{*}$ and the true bounding box of the part {$\Lambda_{I,U^{*}}$}. The CNN is pre-trained to classify object images in ${\bf I}$ of a category from random images. Given the part annotations, \cite{CNNAoG,explanatoryGraph_arXiv} mine \textit{middle-level patterns} and \textit{explanatory patterns} from conv-layers in the CNN to represent neural activations that are highly related to the part annotations. We use the two patterns to build two types of AOGs.

\subsection{AOG visualization}

We learn up-conv-nets~\cite{FeaVisual} to \textit{roughly}\textcolor{red}{\footnote[5]{Up-conv-nets~\cite{FeaVisual} cannot ensure a ``strict'' correspondence between each CNN unit and its visualized appearance.}} visualize the content within AOG nodes. Given a feature map of a CNN's conv-layer as input, the up-conv-net was originally proposed to invert the CNN feature map and output the image corresponding to the feature map, namely, image reconstruction. The AOG encodes several part templates of a semantic part, and each part template consists of latent patterns in different conv-layers. Thus, in each step, we visualize and refine latent patterns of a certain part template. We extend the up-conv-net~\cite{FeaVisual} to visualize latent patterns within the sub-AOG under a given part template $U$.

Given a training object sample $I\in{\bf I}^{\textrm{ant}}$ whose part is explained by the part template $U$ in the AOG, we use $U$'s sub-graph to localize the target part based on Eq.~(\ref{eqn:AND}). During the part-localization process, each latent pattern is assigned with a certain CNN unit. Thus, we can use activation responses of the corresponding CNN units to reconstruct the patterns' appearance. We implement the image reconstruction by simply filtering out all unrelated activation responses from the CNN feature map (\emph{i.e.} setting activations to zero) and using the up-conv-net to invert the modified feature map.

Note that latent patterns of each part template are extracted from different conv-layers of the CNN, and they potentially represent local compositional shapes at different scales. As shown in Fig.~\ref{fig:visual}, latent patterns in lower conv-layers usually represent small-scale details (\emph{e.g.} edges and corners), but these patterns are usually not discriminative enough and more likely to be activated by background noises. In contrast, latent patterns in higher conv-layers mainly correspond to large-scale appearances/contexts without encoding much object details, but they usually consistently represent certain parts among different objects.

As a result, we train different up-conv-nets using feature maps of different conv-layers for image reconstruction, in order to avoid detailed object appearance being mixed with large-scale object patterns during image reconstruction. In this way, we apply different human-interaction strategies for latent patterns in different conv-layers (which will be introduced later).

\textbf{Training and using up-conv-nets:}{\verb| |} Given a CNN that is finetuned for a category, we train an up-conv-net for each conv-layer of the CNN. Training samples are the cropped object images of the category. Given the object images, we use the CNN to extract feature maps of the target conv-layer as the input of the up-conv-net. Output of the up-conv-net is pixel-level image-reconstruction results. The up-conv-net is trained based on the loss of the L-2 distance between the original image $I^{*}$ and the reconstructed image $\hat{I}$:
\begin{equation}
Loss=\Vert{I^{*}-\hat{I}}\Vert_{2}^{2}
\end{equation}

When we apply the up-conv-net to AOG visualization, we use the AOG to localize the target part on the given object $I$. Let the part-parsing process localize a latent pattern $V$ at the horizontal coordinate $(x,y)$ in $d$-th conv-slice of the target conv-layer. We consider that neural responses at $(x,y)$ in this conv-slice are related to $V$\textcolor{red}{\footnote[6]{For conv-layers of VGG-16 whose feature map size $\geq56\times56\times256$, we select neural responses within $(x\pm3,y\pm3)$.}}. We set all neural responses unrelated to any latent patterns to zero in the feature map. Then, we use the up-conv-net to invert the modified feature map for image reconstruction.

\subsection{Human interactions}

Based on visualization results of latent patterns, we further use human interactions to remove latent patterns that are not closely related to the target part. The flowchart of human interactions is designed as follows. Given an part template $U$, our method sequentially produces different image-reconstruction results based on $U$'s latent patterns extracted from different conv-layers. Then, given the reconstructed image \emph{w.r.t.} each conv-layer, we require people to annotate image regions that do \textbf{not} contribute to the localization of the target part. Let $\Lambda^{I}$ denote all image area of $I$, and $\Lambda^{\textrm{ant}}\subset\Lambda^{I}$ denote the union of the annotated image regions. A latent pattern $V$ under $U$ is considered not related to the target part and removed from the AOG, if 1) $V$'s parsed image region $\hat{\Lambda}_{V}$ localizes within $\Lambda^{\textrm{ant}}$, and 2) it satisfies
\begin{equation}
\!\!{\sum}_{p\in\Lambda^{\textrm{ant}}}\vert\frac{\partial S_{I}(V|\hat{\Lambda}_{V})}{\partial p}\vert\!>\!{\sum}_{p\in\Lambda^{I}\setminus\Lambda^{\textrm{ant}}}\vert\frac{\partial S_{I}(V|\hat{\Lambda}_{V})}{\partial p}\vert\!\!
\label{eqn:interaction}
\end{equation}
where $p$ denotes the value of a pixel in $I$. $\frac{\partial S_{I}(V|\hat{\Lambda}_{V})}{\partial p}$ is equal to the gradient of pattern score \emph{w.r.t.} pixel $p$ (see \cite{CNNSemanticPart} for details of the gradient computation). Pixels with high gradients usually have high correspondence to pattern $V$. Compared to the above equation, the up-conv-net can only show \textit{rough} image regions of latent patterns.

\textbf{Rules for human interactions}{\verb| |} We believe that the localization of a certain semantic part mainly relies on two types of information: 1) the contextual information (\emph{e.g.} the global pose of the entire object) for rough part localization from a global view, and 2) detailed part appearance for accurate localization. In general, patterns in low conv-layers describe object details, and those in high conv-layers represent contextual information. Therefore, as shown in Fig.~\ref{fig:visual}, for latent patterns in high conv-layers, we mainly remove those localized on the background outside the ``object.'' For latent patterns in low conv-layers, we usually remove those outside ``part bounding boxes.''

\section{Experiments}

\subsection{Implementation details}

We learned the AOG based on a 16-layer VGG network (VGG-16)~\cite{VGG}, which was pre-trained as follows. The VGG-16 was first pre-trained using the 1.3M images in the ImageNet ILSVRC 2012 dataset~\cite{ImageNet} with a loss for 1000-category classification. Then, we further finetuned the VGG-16 using cropped object images in a category to classify target objects from background images. Just as in \cite{CNNAoG}, we selected the last nine conv-layers in the VGG-16 as valid layers, and extracted latent patterns from these conv-layers to build the AOG.

Note that feature maps for Conv-layers 5--7, Conv-layers 8--10, and Conv-layers 11--13 are three $56\!\times\!56\!\times\!256$, three $28\!\times\!28\!\times512$, and three $14\!\times\!14\!\times512$ matrices, respectively. To simplify the system, we merged conv-slices in Conv-layers 5--7, which contained latent patterns, into a new feature map of $56\!\times\!56\!\times\!N_1$ ($N_1$ denotes the number of valid conv-slices in Conv-layers 5--7). We learned a up-conv-net, namely \textit{Net-1}, to visualize latent patterns in the new feature map. Similarly, we learned \textit{Net-2} and \textit{Net-3} for Conv-layers 8--10 and Conv-layers 11--13, respectively. 


\subsection{Datasets}

In order to make a comprehensive evaluation of the proposed method, we chose three benchmark datasets for testing, \emph{i.e.} the Pascal VOC Part dataset~\cite{SemanticPart}, the CUB200-2011 dataset~\cite{CUB200}, and the ILSVRC 2013 DET Animal-Part dataset~\cite{CNNAoG}. Note that as in many previous studies~\cite{SemanticPart,CNNAoG}, we chose animal categories in these three datasets to evaluate part-localization performance, because animals usually contain multiple non-rigid parts, which presents a key challenge for part localization.

\begin{figure}[t]
\centering
\includegraphics[width=\linewidth]{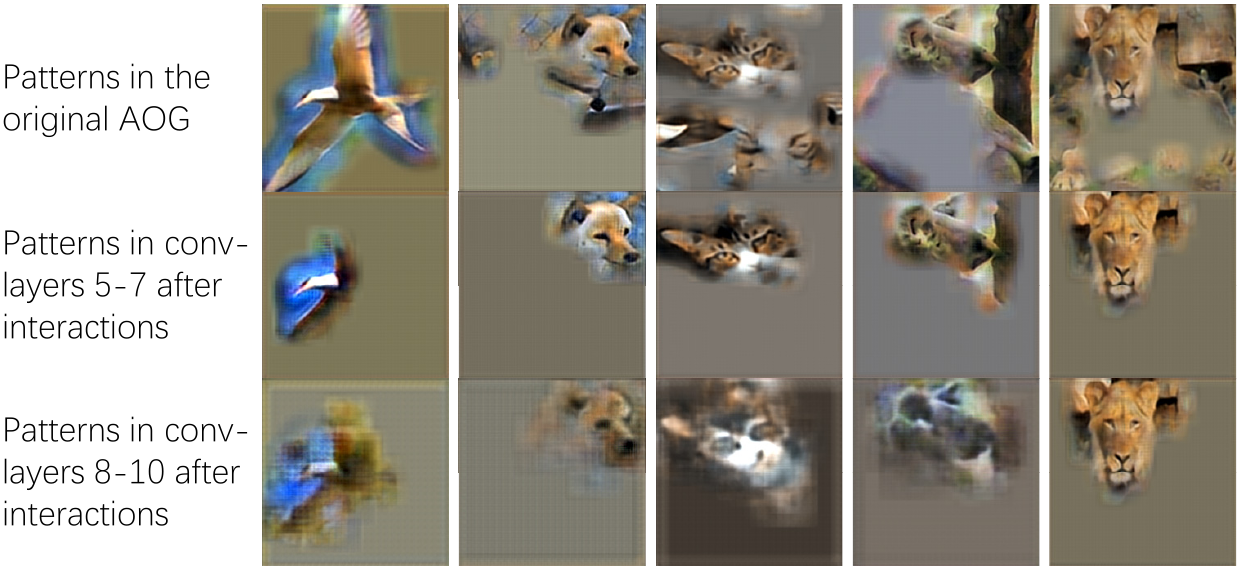}
\caption{Visualization of patterns for the head part before and after human interactions.}
\label{fig:holding}
\end{figure}

\begin{table*}[t]
\centering
\resizebox{1.0\linewidth}{!}{\begin{tabular}{l|c|cccccccccccccccc}
\hline
\multicolumn{2}{r|}{$\qquad\qquad$obj.-box finetune} &\!\!\! gold. \!\!&\!\! bird \!\!&\!\! frog \!\!&\!\! turt. \!\!&\!\! liza. \!\!&\!\! koala \!\!&\!\! lobs. \!\!&\!\! dog \!\!&\!\! fox \!\!&\!\! cat \!\!&\!\! lion \!\!&\!\! tiger \!\!&\!\! bear \!\!&\!\! rabb. \!\!&\!\! hams. \!\!&\!\! squi.\\
\!\!\! {\small SS-DPM-Part~\cite{SSDPM}} \!\!\!&\!\!\! {N}
\!\!\!&\!\!\!{\small0.297}
\!\!\!&\!\!\!{\small0.280}
\!\!\!&\!\!\!{\small0.257}
\!\!\!&\!\!\!{\small0.255}
\!\!\!&\!\!\!{\small0.317}
\!\!\!&\!\!\!{\small0.222}
\!\!\!&\!\!\!{\small0.207}
\!\!\!&\!\!\!{\small0.239}
\!\!\!&\!\!\!{\small0.305}
\!\!\!&\!\!\!{\small0.308}
\!\!\!&\!\!\!{\small0.238}
\!\!\!&\!\!\!{\small0.144}
\!\!\!&\!\!\!{\small0.260}
\!\!\!&\!\!\!{\small0.272}
\!\!\!&\!\!\!{\small0.178}
\!\!\!&\!\!\!{\small0.261}
\\
\!\!\! {\small PL-DPM-Part~\cite{PLDPM}} \!\!\!&\!\!\! {N}
\!\!\!&\!\!\!{\small0.273}
\!\!\!&\!\!\!{\small0.256}
\!\!\!&\!\!\!{\small0.271}
\!\!\!&\!\!\!{\small0.321}
\!\!\!&\!\!\!{\small0.327}
\!\!\!&\!\!\!{\small0.242}
\!\!\!&\!\!\!{\small0.194}
\!\!\!&\!\!\!{\small0.238}
\!\!\!&\!\!\!{\small0.619}
\!\!\!&\!\!\!{\small0.215}
\!\!\!&\!\!\!{\small0.239}
\!\!\!&\!\!\!{\small0.136}
\!\!\!&\!\!\!{\small0.323}
\!\!\!&\!\!\!{\small0.228}
\!\!\!&\!\!\!{\small0.186}
\!\!\!&\!\!\!{\small0.281}
\\
\!\!\! {\small Part-Graph~\cite{SemanticPart}} \!\!\!&\!\!\! {N}
\!\!\!&\!\!\!{\small0.363}
\!\!\!&\!\!\!{\small0.316}
\!\!\!&\!\!\!{\small0.241}
\!\!\!&\!\!\!{\small0.322}
\!\!\!&\!\!\!{\small0.419}
\!\!\!&\!\!\!{\small0.205}
\!\!\!&\!\!\!{\small0.218}
\!\!\!&\!\!\!{\small0.218}
\!\!\!&\!\!\!{\small0.343}
\!\!\!&\!\!\!{\small0.242}
\!\!\!&\!\!\!{\small0.162}
\!\!\!&\!\!\!{\small0.127}
\!\!\!&\!\!\!{\small0.224}
\!\!\!&\!\!\!{\small0.188}
\!\!\!&\!\!\!{\small0.131}
\!\!\!&\!\!\!{\small0.208}
\\
\!\!\! {\small fc7+linearSVM} \!\!\!&\!\!\! {Y}
\!\!\!&\!\!\!{\small0.150}
\!\!\!&\!\!\!{\small0.318}
\!\!\!&\!\!\!{\small0.186}
\!\!\!&\!\!\!{\small0.150}
\!\!\!&\!\!\!{\small0.257}
\!\!\!&\!\!\!{\small0.156}
\!\!\!&\!\!\!{\small0.196}
\!\!\!&\!\!\!{\small0.136}
\!\!\!&\!\!\!{\small0.101}
\!\!\!&\!\!\!{\small0.138}
\!\!\!&\!\!\!{\small0.132}
\!\!\!&\!\!\!{\small0.163}
\!\!\!&\!\!\!{\small0.122}
\!\!\!&\!\!\!{\small0.139}
\!\!\!&\!\!\!{\small0.110}
\!\!\!&\!\!\!{\small0.262}
\\
\!\!\! {\small fc7+RBF-SVM} \!\!\!&\!\!\! {Y}
\!\!\!&\!\!\!{\small0.243}
\!\!\!&\!\!\!{\small0.369}
\!\!\!&\!\!\!{\small0.232}
\!\!\!&\!\!\!{\small0.157}
\!\!\!&\!\!\!{\small0.243}
\!\!\!&\!\!\!{\small0.146}
\!\!\!&\!\!\!{\small0.237}
\!\!\!&\!\!\!{\small0.154}
\!\!\!&\!\!\!{\small0.122}
\!\!\!&\!\!\!{\small0.134}
\!\!\!&\!\!\!{\small0.115}
\!\!\!&\!\!\!{\small0.141}
\!\!\!&\!\!\!{\small0.154}
\!\!\!&\!\!\!{\small0.124}
\!\!\!&\!\!\!{\small0.135}
\!\!\!&\!\!\!{\small0.289}
\\
\!\!\! {\small fc7+NN} \!\!\!&\!\!\! {Y}
\!\!\!&\!\!\!{\small0.298}
\!\!\!&\!\!\!{\small0.387}
\!\!\!&\!\!\!{\small0.307}
\!\!\!&\!\!\!{\small0.259}
\!\!\!&\!\!\!{\small0.300}
\!\!\!&\!\!\!{\small0.169}
\!\!\!&\!\!\!{\small0.287}
\!\!\!&\!\!\!{\small0.242}
\!\!\!&\!\!\!{\small0.170}
\!\!\!&\!\!\!{\small0.159}
\!\!\!&\!\!\!{\small0.112}
\!\!\!&\!\!\!{\small0.135}
\!\!\!&\!\!\!{\small0.263}
\!\!\!&\!\!\!{\small0.219}
\!\!\!&\!\!\!{\small0.152}
\!\!\!&\!\!\!{\small0.346}
\\
\!\!\! {\small fc7+sp+linearSVM} \!\!\!&\!\!\! {Y}
\!\!\!&\!\!\!{0.150}
\!\!\!&\!\!\!{0.318}
\!\!\!&\!\!\!{0.186}
\!\!\!&\!\!\!{0.150}
\!\!\!&\!\!\!{0.254}
\!\!\!&\!\!\!{0.156}
\!\!\!&\!\!\!{0.196}
\!\!\!&\!\!\!{0.136}
\!\!\!&\!\!\!{0.101}
\!\!\!&\!\!\!{0.138}
\!\!\!&\!\!\!{0.132}
\!\!\!&\!\!\!{0.163}
\!\!\!&\!\!\!{0.122}
\!\!\!&\!\!\!{0.139}
\!\!\!&\!\!\!{0.110}
\!\!\!&\!\!\!{0.262}
\\
\!\!\! {\small fc7+sp+RBF-SVM} \!\!\!&\!\!\! {Y}
\!\!\!&\!\!\!{0.243}
\!\!\!&\!\!\!{0.371}
\!\!\!&\!\!\!{0.235}
\!\!\!&\!\!\!{0.156}
\!\!\!&\!\!\!{0.252}
\!\!\!&\!\!\!{0.145}
\!\!\!&\!\!\!{0.237}
\!\!\!&\!\!\!{0.154}
\!\!\!&\!\!\!{0.122}
\!\!\!&\!\!\!{0.134}
\!\!\!&\!\!\!{0.115}
\!\!\!&\!\!\!{0.140}
\!\!\!&\!\!\!{0.156}
\!\!\!&\!\!\!{0.121}
\!\!\!&\!\!\!{0.136}
\!\!\!&\!\!\!{0.302}
\\
\!\!\! {\small fc7+sp+NN} \!\!\!&\!\!\! {Y}
\!\!\!&\!\!\!{0.298}
\!\!\!&\!\!\!{0.387}
\!\!\!&\!\!\!{0.307}
\!\!\!&\!\!\!{0.259}
\!\!\!&\!\!\!{0.300}
\!\!\!&\!\!\!{0.169}
\!\!\!&\!\!\!{0.287}
\!\!\!&\!\!\!{0.242}
\!\!\!&\!\!\!{0.170}
\!\!\!&\!\!\!{0.158}
\!\!\!&\!\!\!{0.112}
\!\!\!&\!\!\!{0.135}
\!\!\!&\!\!\!{0.263}
\!\!\!&\!\!\!{0.219}
\!\!\!&\!\!\!{0.152}
\!\!\!&\!\!\!{0.345}
\\
\!\!\! {\small CNN-PDD~\cite{CNNSemanticPart}} \!\!\!&\!\!\! {N}
\!\!\!&\!\!\!{\small0.316}
\!\!\!&\!\!\!{\small0.289}
\!\!\!&\!\!\!{\small0.229}
\!\!\!&\!\!\!{\small0.260}
\!\!\!&\!\!\!{\small0.335}
\!\!\!&\!\!\!{\small0.163}
\!\!\!&\!\!\!{\small0.190}
\!\!\!&\!\!\!{\small0.220}
\!\!\!&\!\!\!{\small0.212}
\!\!\!&\!\!\!{\small0.196}
\!\!\!&\!\!\!{\small0.174}
\!\!\!&\!\!\!{\small0.160}
\!\!\!&\!\!\!{\small0.223}
\!\!\!&\!\!\!{\small0.266}
\!\!\!&\!\!\!{\small0.156}
\!\!\!&\!\!\!{\small0.291}
\\
\!\!\! {\small CNN-PDD-ft~\cite{CNNSemanticPart}} \!\!\!&\!\!\! {Y}
\!\!\!&\!\!\!{\small0.302}
\!\!\!&\!\!\!{\small0.236}
\!\!\!&\!\!\!{\small0.261}
\!\!\!&\!\!\!{\small0.231}
\!\!\!&\!\!\!{\small0.350}
\!\!\!&\!\!\!{\small0.168}
\!\!\!&\!\!\!{\small0.170}
\!\!\!&\!\!\!{\small0.177}
\!\!\!&\!\!\!{\small0.264}
\!\!\!&\!\!\!{\small0.270}
\!\!\!&\!\!\!{\small0.206}
\!\!\!&\!\!\!{\small0.256}
\!\!\!&\!\!\!{\small0.178}
\!\!\!&\!\!\!{\small0.167}
\!\!\!&\!\!\!{\small0.286}
\!\!\!&\!\!\!{\small0.237}
\\
\!\!\! {\small Fast-RCNN (1 ft)~\cite{FastRCNN}} \!\!\!&\!\!\! {N}
\!\!\!&\!\!\!{\small0.313}
\!\!\!&\!\!\!{\small0.370}
\!\!\!&\!\!\!{\small0.250}
\!\!\!&\!\!\!{\small0.318}
\!\!\!&\!\!\!{\small0.375}
\!\!\!&\!\!\!{\small0.343}
\!\!\!&\!\!\!{\small0.291}
\!\!\!&\!\!\!{\small0.365}
\!\!\!&\!\!\!{\small0.287}
\!\!\!&\!\!\!{\small0.321}
\!\!\!&\!\!\!{\small0.291}
\!\!\!&\!\!\!{\small0.305}
\!\!\!&\!\!\!{\small0.349}
\!\!\!&\!\!\!{\small0.261}
\!\!\!&\!\!\!{\small0.290}
\!\!\!&\!\!\!{\small0.165}
\\
\!\!\! {\small Fast-RCNN (2 fts)~\cite{FastRCNN}} \!\!\!&\!\!\! {Y}
\!\!\!&\!\!\!{\small0.355}
\!\!\!&\!\!\!{\small0.382}
\!\!\!&\!\!\!{\small0.421}
\!\!\!&\!\!\!{\small0.298}
\!\!\!&\!\!\!{\small0.413}
\!\!\!&\!\!\!{\small0.168}
\!\!\!&\!\!\!{\small0.328}
\!\!\!&\!\!\!{\small0.383}
\!\!\!&\!\!\!{\small0.298}
\!\!\!&\!\!\!{\small0.219}
\!\!\!&\!\!\!{\small0.196}
\!\!\!&\!\!\!{\small0.217}
\!\!\!&\!\!\!{\small0.245}
\!\!\!&\!\!\!{\small0.265}
\!\!\!&\!\!\!{\small0.264}
\!\!\!&\!\!\!{\small0.220}
\\
\!\!\! {\small Mining-raw~\cite{CNNAoG}} \!\!\!&\!\!\! {Y}
\!\!\!&\!\!\!{\small0.102}
\!\!\!&\!\!\!{\small0.149}
\!\!\!&\!\!\!{\small0.116}
\!\!\!&\!\!\!{\small0.215}
\!\!\!&\!\!\!{\small0.137}
\!\!\!&\!\!\!{\small0.094}
\!\!\!&\!\!\!{\small0.162}
\!\!\!&\!\!\!{\small0.146}
\!\!\!&\!\!\!{\small0.081}
\!\!\!&\!\!\!{\small0.154}
\!\!\!&\!\!\!{\small0.079}
\!\!\!&\!\!\!{\small0.088}
\!\!\!&\!\!\!{\small0.120}
\!\!\!&\!\!\!{\small0.092}
\!\!\!&\!\!\!{\small0.094}
\!\!\!&\!\!\!{\small0.105}
\\
\!\!\! {\small Ours} \!\!\!&\!\!\! {Y}
\!\!\!&\!\!\!{\small\bf 0.074}
\!\!\!&\!\!\!{\small\bf 0.108}
\!\!\!&\!\!\!{\small\bf 0.095}
\!\!\!&\!\!\!{\small\bf 0.182}
\!\!\!&\!\!\!{\small\bf 0.129}
\!\!\!&\!\!\!{\small\bf 0.079}
\!\!\!&\!\!\!{\small\bf 0.147}
\!\!\!&\!\!\!{\small\bf 0.124}
\!\!\!&\!\!\!{\small\bf 0.058}
\!\!\!&\!\!\!{\small\bf 0.132}
\!\!\!&\!\!\!{\small\bf 0.071}
\!\!\!&\!\!\!{\small\bf 0.083}
\!\!\!&\!\!\!{\small\bf 0.104}
\!\!\!&\!\!\!{\small\bf 0.078}
\!\!\!&\!\!\!{\small\bf 0.077}
\!\!\!&\!\!\!{\small\bf 0.072}
\\
\hline
\!\!\!&\!\!\! \!\!\!&\!\!\! horse \!\!\!&\!\!\! zebra \!\!\!&\!\!\! swine \!\!\!&\!\!\! hippo \!\!\!&\!\!\! catt. \!\!\!&\!\!\! sheep \!\!\!&\!\!\! ante. \!\!\!&\!\!\! camel \!\!\!&\!\!\! otter \!\!\!&\!\!\! arma. \!\!\!&\!\!\! monk. \!\!\!&\!\!\! elep. \!\!\!&\!\!\! red pa. \!\!\!&\!\!\! gia.pa. \!\!\!&\!\!\! \!\!\!&\!\!\! \textcolor{blue}{\bf\large Avg.}\\
\!\!\! {\small SS-DPM-Part~\cite{SSDPM}} \!\!\!&\!\!\! {N}
\!\!\!&\!\!\!{\small0.246}
\!\!\!&\!\!\!{\small0.206}
\!\!\!&\!\!\!{\small0.240}
\!\!\!&\!\!\!{\small0.234}
\!\!\!&\!\!\!{\small0.246}
\!\!\!&\!\!\!{\small0.205}
\!\!\!&\!\!\!{\small0.224}
\!\!\!&\!\!\!{\small0.277}
\!\!\!&\!\!\!{\small0.253}
\!\!\!&\!\!\!{\small0.283}
\!\!\!&\!\!\!{\small0.206}
\!\!\!&\!\!\!{\small0.219}
\!\!\!&\!\!\!{\small0.256}
\!\!\!&\!\!\!{\small0.129}
\!\!\!&\!\!\!
\!\!\!&\!\!\!{\small\textcolor{blue}{0.242}}
\\
\!\!\! {\small PL-DPM-Part~\cite{PLDPM}} \!\!\!&\!\!\! {N}
\!\!\!&\!\!\!{\small0.322}
\!\!\!&\!\!\!{\small0.267}
\!\!\!&\!\!\!{\small0.297}
\!\!\!&\!\!\!{\small0.273}
\!\!\!&\!\!\!{\small0.271}
\!\!\!&\!\!\!{\small0.413}
\!\!\!&\!\!\!{\small0.337}
\!\!\!&\!\!\!{\small0.261}
\!\!\!&\!\!\!{\small0.286}
\!\!\!&\!\!\!{\small0.295}
\!\!\!&\!\!\!{\small0.187}
\!\!\!&\!\!\!{\small0.264}
\!\!\!&\!\!\!{\small0.204}
\!\!\!&\!\!\!{\small0.505}
\!\!\!&\!\!\!
\!\!\!&\!\!\!{\small\textcolor{blue}{0.284}}
\\
\!\!\! {\small Part-Graph~\cite{SemanticPart}} \!\!\!&\!\!\! {N}
\!\!\!&\!\!\!{\small0.296}
\!\!\!&\!\!\!{\small0.315}
\!\!\!&\!\!\!{\small0.306}
\!\!\!&\!\!\!{\small0.378}
\!\!\!&\!\!\!{\small0.333}
\!\!\!&\!\!\!{\small0.230}
\!\!\!&\!\!\!{\small0.216}
\!\!\!&\!\!\!{\small0.317}
\!\!\!&\!\!\!{\small0.227}
\!\!\!&\!\!\!{\small0.341}
\!\!\!&\!\!\!{\small0.159}
\!\!\!&\!\!\!{\small0.294}
\!\!\!&\!\!\!{\small0.276}
\!\!\!&\!\!\!{\small0.094}
\!\!\!&\!\!\!
\!\!\!&\!\!\!{\small\textcolor{blue}{0.257}}
\\
\!\!\! {\small fc7+linearSVM} \!\!\!&\!\!\! {Y}
\!\!\!&\!\!\!{\small0.205}
\!\!\!&\!\!\!{\small0.258}
\!\!\!&\!\!\!{\small0.201}
\!\!\!&\!\!\!{\small0.140}
\!\!\!&\!\!\!{\small0.256}
\!\!\!&\!\!\!{\small0.236}
\!\!\!&\!\!\!{\small0.164}
\!\!\!&\!\!\!{\small0.190}
\!\!\!&\!\!\!{\small0.140}
\!\!\!&\!\!\!{\small0.252}
\!\!\!&\!\!\!{\small0.256}
\!\!\!&\!\!\!{\small0.176}
\!\!\!&\!\!\!{\small0.215}
\!\!\!&\!\!\!{\small0.116}
\!\!\!&\!\!\!
\!\!\!&\!\!\!{\small\textcolor{blue}{0.184}}
\\
\!\!\! {\small fc7+RBF-SVM} \!\!\!&\!\!\! {Y}
\!\!\!&\!\!\!{\small0.234}
\!\!\!&\!\!\!{\small0.221}
\!\!\!&\!\!\!{\small0.237}
\!\!\!&\!\!\!{\small0.168}
\!\!\!&\!\!\!{\small0.300}
\!\!\!&\!\!\!{\small0.253}
\!\!\!&\!\!\!{\small0.171}
\!\!\!&\!\!\!{\small0.212}
\!\!\!&\!\!\!{\small0.146}
\!\!\!&\!\!\!{\small0.238}
\!\!\!&\!\!\!{\small0.248}
\!\!\!&\!\!\!{\small0.225}
\!\!\!&\!\!\!{\small0.185}
\!\!\!&\!\!\!{\small0.104}
\!\!\!&\!\!\!
\!\!\!&\!\!\!{\small\textcolor{blue}{0.198}}
\\
\!\!\! {\small fc7+NN} \!\!\!&\!\!\! {Y}
\!\!\!&\!\!\!{\small0.293}
\!\!\!&\!\!\!{\small0.235}
\!\!\!&\!\!\!{\small0.297}
\!\!\!&\!\!\!{\small0.335}
\!\!\!&\!\!\!{\small0.330}
\!\!\!&\!\!\!{\small0.262}
\!\!\!&\!\!\!{\small0.305}
\!\!\!&\!\!\!{\small0.263}
\!\!\!&\!\!\!{\small0.125}
\!\!\!&\!\!\!{\small0.262}
\!\!\!&\!\!\!{\small0.304}
\!\!\!&\!\!\!{\small0.277}
\!\!\!&\!\!\!{\small0.214}
\!\!\!&\!\!\!{\small0.102}
\!\!\!&\!\!\!
\!\!\!&\!\!\!{\small\textcolor{blue}{0.247}}
\\
\!\!\! {\small fc7+sp+linearSVM} \!\!\!&\!\!\! {Y}
\!\!\!&\!\!\!{0.205}
\!\!\!&\!\!\!{0.258}
\!\!\!&\!\!\!{0.201}
\!\!\!&\!\!\!{\bf 0.140}
\!\!\!&\!\!\!{0.256}
\!\!\!&\!\!\!{0.236}
\!\!\!&\!\!\!{0.164}
\!\!\!&\!\!\!{0.190}
\!\!\!&\!\!\!{0.140}
\!\!\!&\!\!\!{0.250}
\!\!\!&\!\!\!{0.256}
\!\!\!&\!\!\!{0.176}
\!\!\!&\!\!\!{0.215}
\!\!\!&\!\!\!{0.116}
\!\!\!&\!\!\!
\!\!\!&\!\!\!{\textcolor{blue}{0.184}}
\\
\!\!\! {\small fc7+sp+RBF-SVM} \!\!\!&\!\!\! {Y}
\!\!\!&\!\!\!{0.234}
\!\!\!&\!\!\!{0.221}
\!\!\!&\!\!\!{0.237}
\!\!\!&\!\!\!{0.165}
\!\!\!&\!\!\!{0.290}
\!\!\!&\!\!\!{0.246}
\!\!\!&\!\!\!{0.181}
\!\!\!&\!\!\!{0.211}
\!\!\!&\!\!\!{0.146}
\!\!\!&\!\!\!{0.238}
\!\!\!&\!\!\!{0.250}
\!\!\!&\!\!\!{0.224}
\!\!\!&\!\!\!{0.184}
\!\!\!&\!\!\!{0.103}
\!\!\!&\!\!\!
\!\!\!&\!\!\!{\textcolor{blue}{0.198}}
\\
\!\!\! {\small fc7+sp+NN} \!\!\!&\!\!\! {Y}
\!\!\!&\!\!\!{0.293}
\!\!\!&\!\!\!{0.235}
\!\!\!&\!\!\!{0.299}
\!\!\!&\!\!\!{0.335}
\!\!\!&\!\!\!{0.330}
\!\!\!&\!\!\!{0.262}
\!\!\!&\!\!\!{0.305}
\!\!\!&\!\!\!{0.262}
\!\!\!&\!\!\!{0.125}
\!\!\!&\!\!\!{0.260}
\!\!\!&\!\!\!{0.304}
\!\!\!&\!\!\!{0.276}
\!\!\!&\!\!\!{0.214}
\!\!\!&\!\!\!{0.102}
\!\!\!&\!\!\!
\!\!\!&\!\!\!{\textcolor{blue}{0.247}}
\\
\!\!\! {\small CNN-PDD~\cite{CNNSemanticPart}} \!\!\!&\!\!\! {N}
\!\!\!&\!\!\!{\small0.261}
\!\!\!&\!\!\!{\small0.266}
\!\!\!&\!\!\!{\small\bf 0.189}
\!\!\!&\!\!\!{\small0.192}
\!\!\!&\!\!\!{\small0.201}
\!\!\!&\!\!\!{\small0.244}
\!\!\!&\!\!\!{\small0.208}
\!\!\!&\!\!\!{\small0.193}
\!\!\!&\!\!\!{\small0.174}
\!\!\!&\!\!\!{\small0.299}
\!\!\!&\!\!\!{\small0.236}
\!\!\!&\!\!\!{\small0.214}
\!\!\!&\!\!\!{\small0.222}
\!\!\!&\!\!\!{\small0.179}
\!\!\!&\!\!\!
\!\!\!&\!\!\!{\small\textcolor{blue}{0.225}}
\\
\!\!\! {\small CNN-PDD-ft~\cite{CNNSemanticPart}} \!\!\!&\!\!\! {Y}
\!\!\!&\!\!\!{\small0.310}
\!\!\!&\!\!\!{\small0.321}
\!\!\!&\!\!\!{\small0.216}
\!\!\!&\!\!\!{\small0.257}
\!\!\!&\!\!\!{\small0.220}
\!\!\!&\!\!\!{\small0.179}
\!\!\!&\!\!\!{\small0.229}
\!\!\!&\!\!\!{\small0.253}
\!\!\!&\!\!\!{\small0.198}
\!\!\!&\!\!\!{\small0.308}
\!\!\!&\!\!\!{\small0.273}
\!\!\!&\!\!\!{\small0.189}
\!\!\!&\!\!\!{\small0.208}
\!\!\!&\!\!\!{\small0.275}
\!\!\!&\!\!\!
\!\!\!&\!\!\!{\small\textcolor{blue}{0.240}}
\\
\!\!\! {\small Fast-RCNN (1 ft)~\cite{FastRCNN}} \!\!\!&\!\!\! {N}
\!\!\!&\!\!\!{\small0.372}
\!\!\!&\!\!\!{\small0.360}
\!\!\!&\!\!\!{\small0.302}
\!\!\!&\!\!\!{\small0.289}
\!\!\!&\!\!\!{\small0.342}
\!\!\!&\!\!\!{\small0.266}
\!\!\!&\!\!\!{\small0.220}
\!\!\!&\!\!\!{\small0.349}
\!\!\!&\!\!\!{\small0.328}
\!\!\!&\!\!\!{\small0.334}
\!\!\!&\!\!\!{\small0.351}
\!\!\!&\!\!\!{\small0.261}
\!\!\!&\!\!\!{\small0.337}
\!\!\!&\!\!\!{\small0.328}
\!\!\!&\!\!\!
\!\!\!&\!\!\!{\small\textcolor{blue}{0.311}}
\\
\!\!\! {\small Fast-RCNN (2 fts)~\cite{FastRCNN}} \!\!\!&\!\!\! {Y}
\!\!\!&\!\!\!{\small0.355}
\!\!\!&\!\!\!{\small0.324}
\!\!\!&\!\!\!{\small0.275}
\!\!\!&\!\!\!{\small0.266}
\!\!\!&\!\!\!{\small0.292}
\!\!\!&\!\!\!{\small0.235}
\!\!\!&\!\!\!{\small0.212}
\!\!\!&\!\!\!{\small0.271}
\!\!\!&\!\!\!{\small0.329}
\!\!\!&\!\!\!{\small0.343}
\!\!\!&\!\!\!{\small0.259}
\!\!\!&\!\!\!{\small0.163}
\!\!\!&\!\!\!{\small0.285}
\!\!\!&\!\!\!{\small0.246}
\!\!\!&\!\!\!
\!\!\!&\!\!\!{\small\textcolor{blue}{0.284}}
\\
\!\!\! {\small Mining-raw~\cite{CNNAoG}} \!\!\!&\!\!\! {Y}
\!\!\!&\!\!\!{\small0.182}
\!\!\!&\!\!\!{\small0.160}
\!\!\!&\!\!\!{\small0.211}
\!\!\!&\!\!\!{\small0.169}
\!\!\!&\!\!\!{\small0.186}
\!\!\!&\!\!\!{\small0.117}
\!\!\!&\!\!\!{\small0.102}
\!\!\!&\!\!\!{\small0.145}
\!\!\!&\!\!\!{\small0.117}
\!\!\!&\!\!\!{\small0.241}
\!\!\!&\!\!\!{\small0.113}
\!\!\!&\!\!\!{\small0.141}
\!\!\!&\!\!\!{\small0.135}
\!\!\!&\!\!\!{\small0.081}
\!\!\!&\!\!\!
\!\!\!&\!\!\!{\small\textcolor{blue}{0.135}}
\\
\!\!\! {\small Ours} \!\!\!&\!\!\! {Y}
\!\!\!&\!\!\!{\small\bf 0.158}
\!\!\!&\!\!\!{\small\bf 0.126}
\!\!\!&\!\!\!{\small0.196}
\!\!\!&\!\!\!{\small0.161}
\!\!\!&\!\!\!{\small\bf 0.155}
\!\!\!&\!\!\!{\small\bf 0.113}
\!\!\!&\!\!\!{\small\bf 0.092}
\!\!\!&\!\!\!{\small\bf 0.120}
\!\!\!&\!\!\!{\small\bf 0.097}
\!\!\!&\!\!\!{\small\bf 0.216}
\!\!\!&\!\!\!{\small\bf 0.081}
\!\!\!&\!\!\!{\small\bf 0.138}
\!\!\!&\!\!\!{\small\bf 0.109}
\!\!\!&\!\!\!{\small\bf 0.074}
\!\!\!&\!\!\!
\!\!\!&\!\!\!{\small\textcolor{blue}{\bf 0.115}}
\\
\hline
\end{tabular}}
\vspace{2pt}
\caption{Normalized distance of part localization on the ILSVRC 2013 DET Animal-Part dataset. The second column indicates whether the baseline used all object annotations in the category to pre-finetune a CNN before learning the part (in fact, \textit{object-box annotations are more than part annotations}).}
\label{tab:imgnet}
\end{table*}

\subsection{Baselines}

In experiments, we used our interactive learning method to refine the AOG learned by \cite{CNNAoG}, namely the \textit{Ours} method. Then, we followed the algorithm of \cite{CNNAoG} to further build a new AOG, which used \textit{explanatory patterns} in \cite{explanatoryGraph_arXiv} as latent patterns. We conducted the interactive learning on the new AOG, and named it the \textit{Ours (explanatory patterns)} method.

We compared our methods with a total of fourteen methods in part localization. The baselines included 1) state-of-the-art algorithms for object detection (\emph{i.e.} directly detecting target parts from objects), 2) conventional graphical/part models for part localization (modeling both the part appearance and the relationships with parts and objects), and 3) the methods selecting CNN patterns to describe object parts.

The first baseline was the standard method of using AOGs for part localization without applying human interactions~\cite{CNNAoG}, namely \textit{Mining-raw}. The comparison with \textit{Mining-raw} demonstrated the effectiveness of human interactions in learning.

Then, two baselines were implemented based on the Fast-RCNN~\cite{FastRCNN}. We finetuned the fast-RCNN with a loss for detecting a single class/part from background, rather than for multi-class/part detection to ensure a fair comparison. The first baseline was the standard fast-RCNN, namely \textit{Fast-RCNN (1 ft)}, which directly finetuned the VGG-16 network to detect parts on objects. We annotated object parts on well cropped object images as training samples (objects had been cropped using their bounding boxes). Then, for the second baseline, namely \textit{Fast-RCNN (2 fts)}, we slightly modified original algorithm of Fast-RCNN. Given the large number of \textbf{object}-box annotations in the target category, the Fast-RCNN (2 fts) first finetuned the VGG-16 network with the loss of ``detecting objects from entire images.'' Then, given a small number of \textbf{part} annotations, Fast-RCNN (2 fts) further finetuned the VGG-16 to detect parts from objects. This two-step finetuning made a full use of the massive object-level annotations, which enabled a fair comparison.

\begin{table}[t]
\centering
\resizebox{0.8\linewidth}{!}{\begin{tabular}{lcc}
\hline
Method & Finetune &\\
\hline
{SS-DPM-Part}~\cite{SSDPM} & N &0.3469\\
{PL-DPM-Part}~\cite{PLDPM} & N &0.3412\\
{Part-Graph}~\cite{SemanticPart} & N &0.4889\\
{fc7+linearSVM} & Y &0.3120\\
{fc7+RBF-SVM} & Y &0.3666\\
{fc7+NN} & Y &0.4194\\
{fc7+sp+linearSVM} & Y &0.3120\\
{fc7+sp+RBF-SVM} & Y &0.3700\\
{fc7+sp+NN} & Y &0.4195\\
{CNN-PDD}~\cite{CNNSemanticPart} & N &0.2333\\
{CNN-PDD-ft}~\cite{CNNSemanticPart} & Y &0.3269\\
{Fast-RCNN (1 ft)}~\cite{FastRCNN} & N &0.4517\\
{Fast-RCNN (2 fts)}~\cite{FastRCNN} & Y &0.4131\\
{Mining-raw}~\cite{CNNAoG} & Y &0.0915\\
{Ours} & Y &{0.0878}\\
{Ours (explanatory patterns)} & Y &{\bf0.0860}\\
\hline
\end{tabular}}
\vspace{2pt}
\caption{Normalized distance of part localization on the CUB200-2011 dataset. The second column indicates whether the baseline used all object annotations in the category to pre-finetune a CNN before learning the part. 
}
\label{tab:cub200}
\end{table}

\begin{table}[t]
\centering
\resizebox{1.0\linewidth}{!}{\begin{tabular}{l|ccccccc}
\hline
&bird$\;$ \!\!\!\!&\!\!\!\! $\;$cat$\;$ \!\!\!\!&\!\!\!\! $\;$cow$\;$ \!\!\!\!&\!\!\!\! $\;$dog$\;$ \!\!\!\!&\!\!\!\! $\;$horse \!\!\!\!&\!\!\!\! sheep \!\!\!\!&\!\!\!\! $\;$\textcolor{blue}{\textbf{Avg}}\\
\hline
\!\!\! {\footnotesize SS-DPM-Part}\textcolor{white}{A}\!\!\!\!\!
\!\!&\!{\footnotesize0.356}
\!\!&\!{\footnotesize0.270}
\!\!&\!{\footnotesize0.264}
\!\!&\!{\footnotesize0.242}
\!\!&\!{\footnotesize0.262}
\!\!&\!{\footnotesize0.286}
\!\!&\!\textcolor{blue}{\footnotesize0.280}
\\
\!\!\! {\footnotesize PL-DPM-Part}\textcolor{white}{A}\!\!\!
\!\!&\!{\footnotesize0.294}
\!\!&\!{\footnotesize0.328}
\!\!&\!{\footnotesize0.282}
\!\!&\!{\footnotesize0.312}
\!\!&\!{\footnotesize0.321}
\!\!&\!{\footnotesize0.840}
\!\!&\!\textcolor{blue}{\footnotesize0.396}
\\
\!\!\! {\footnotesize Part-Graph}\textcolor{white}{A}\!\!\!
\!\!&\!{\footnotesize0.360}
\!\!&\!{\footnotesize0.208}
\!\!&\!{\footnotesize0.263}
\!\!&\!{\footnotesize0.205}
\!\!&\!{\footnotesize0.386}
\!\!&\!{\footnotesize0.500}
\!\!&\!\textcolor{blue}{\footnotesize0.320}
\\
\!\!\! {\footnotesize fc7+linearSVM}\textcolor{white}{A}\!\!\!
\!\!&\!{\footnotesize0.247}
\!\!&\!{\footnotesize0.174}
\!\!&\!{\footnotesize0.251}
\!\!&\!{\footnotesize0.217}
\!\!&\!{\footnotesize0.261}
\!\!&\!{\footnotesize0.317}
\!\!&\!\textcolor{blue}{\footnotesize0.244}
\\
\!\!\! {\footnotesize fc7+RBF-SVM}\textcolor{white}{A}\!\!\!
\!\!&\!{\footnotesize0.276}
\!\!&\!{\footnotesize0.167}
\!\!&\!{\footnotesize0.265}
\!\!&\!{\footnotesize0.244}
\!\!&\!{\footnotesize0.263}
\!\!&\!{\footnotesize0.313}
\!\!&\!\textcolor{blue}{\footnotesize0.255}
\\
\!\!\! {\footnotesize fc7+NN}\textcolor{white}{A}\!\!\!
\!\!&\!{\footnotesize0.344}
\!\!&\!{\footnotesize0.188}
\!\!&\!{\footnotesize0.316}
\!\!&\!{\footnotesize0.288}
\!\!&\!{\footnotesize0.351}
\!\!&\!{\footnotesize0.313}
\!\!&\!\textcolor{blue}{\footnotesize0.300}
\\
\!\!\! {\footnotesize fc7+sp+linearSVM}\textcolor{white}{A}\!\!\!
\!\!&\!{\footnotesize0.247}
\!\!&\!{\footnotesize0.174}
\!\!&\!{\footnotesize0.249}
\!\!&\!{\footnotesize0.217}
\!\!&\!{\footnotesize0.261}
\!\!&\!{\footnotesize0.317}
\!\!&\!\textcolor{blue}{\footnotesize0.244}
\\
\!\!\! {\footnotesize fc7+sp+RBF-SVM}\textcolor{white}{A}\!\!\!
\!\!&\!{\footnotesize0.276}
\!\!&\!{\footnotesize0.167}
\!\!&\!{\footnotesize0.273}
\!\!&\!{\footnotesize0.242}
\!\!&\!{\footnotesize0.264}
\!\!&\!{\footnotesize0.309}
\!\!&\!\textcolor{blue}{\footnotesize0.255}
\\
\!\!\! {\footnotesize fc7+sp+NN}\textcolor{white}{A}\!\!\!
\!\!&\!{\footnotesize0.344}
\!\!&\!{\footnotesize0.188}
\!\!&\!{\footnotesize0.316}
\!\!&\!{\footnotesize0.288}
\!\!&\!{\footnotesize0.351}
\!\!&\!{\footnotesize0.313}
\!\!&\!\textcolor{blue}{\footnotesize0.300}
\\
\!\!\! {\footnotesize CNN-PDD}\textcolor{white}{A}\!\!\!
\!\!&\!{\footnotesize0.301}
\!\!&\!{\footnotesize0.246}
\!\!&\!{\footnotesize0.220}
\!\!&\!{\footnotesize0.248}
\!\!&\!{\footnotesize0.292}
\!\!&\!{\footnotesize0.254}
\!\!&\!\textcolor{blue}{\footnotesize0.260}
\\
\!\!\! {\footnotesize CNN-PDD-ft}\textcolor{white}{A}\!\!\!
\!\!&\!{\footnotesize0.358}
\!\!&\!{\footnotesize0.268}
\!\!&\!{\footnotesize0.220}
\!\!&\!{\footnotesize0.200}
\!\!&\!{\footnotesize0.302}
\!\!&\!{\footnotesize0.269}
\!\!&\!\textcolor{blue}{\footnotesize0.269}
\\
\!\!\! {\footnotesize Fast-RCNN (1 ft)}\textcolor{white}{A}\!\!\!
\!\!&\!{\footnotesize0.324}
\!\!&\!{\footnotesize0.324}
\!\!&\!{\footnotesize0.325}
\!\!&\!{\footnotesize0.272}
\!\!&\!{\footnotesize0.347}
\!\!&\!{\footnotesize0.314}
\!\!&\!\textcolor{blue}{\footnotesize0.318}
\\
\!\!\! {\footnotesize Fast-RCNN (2 fts)}\textcolor{white}{A}\!\!\!
\!\!&\!{\footnotesize0.350}
\!\!&\!{\footnotesize0.295}
\!\!&\!{\footnotesize0.255}
\!\!&\!{\footnotesize0.293}
\!\!&\!{\footnotesize0.367}
\!\!&\!{\footnotesize0.260}
\!\!&\!\textcolor{blue}{\footnotesize0.303}
\\
\!\!\! {\footnotesize Mining-raw}\textcolor{white}{A}\!\!\!
\!\!&\!{\footnotesize0.187}
\!\!&\!{\footnotesize0.132}
\!\!&\!{\footnotesize0.212}
\!\!&\!{\footnotesize0.175}
\!\!&\!{\footnotesize0.168}
\!\!&\!{\footnotesize0.186}
\!\!&\!\textcolor{blue}{\footnotesize0.177}
\\
\!\!\! {\footnotesize Ours \scriptsize(explanatory patterns)}\textcolor{white}{A}\!\!\!
\!\!&\!{\footnotesize0.162}
\!\!&\!{\footnotesize0.128}
\!\!&\!{\footnotesize0.258}
\!\!&\!{\footnotesize\bf0.137}
\!\!&\!{\footnotesize0.179}
\!\!&\!{\footnotesize0.187}
\!\!&\!\textcolor{blue}{\footnotesize0.175}
\\
\!\!\! {\footnotesize Ours}\textcolor{white}{A}\!\!\!
\!\!&\!{\footnotesize\bf0.129}
\!\!&\!{\footnotesize\bf0.108}
\!\!&\!{\footnotesize\bf0.184}
\!\!&\!{\footnotesize0.140}
\!\!&\!{\footnotesize\bf0.160}
\!\!&\!{\footnotesize\bf0.156}
\!\!&\!\textcolor{blue}{\footnotesize\bf0.146}
\\
\hline
\end{tabular}}
\vspace{2pt}
\caption{Normalized distance of part localization on the Pascal VOC Part dataset.}
\label{tab:voc}
\end{table}

Another typical baseline was \textit{CNN-PDD} proposed in \cite{CNNSemanticPart}. CNN-PDD selected certain conv-slices (channels) of a CNN to represent the target part for part localization. In CNN-PDD, the CNN was pre-trained using 1.3M images in the ImageNet ILSVRC 2012 dataset. Just like Fast-RCNN (1 ft), we also extended the method of \cite{CNNSemanticPart} as a new baseline \textit{CNN-PDD-ft}, in order to make a full use of object-level annotations. \textit{CNN-PDD-ft} pre-finetuned the VGG-16 network using object-box annotations of the target category before the selection of CNN channels.

\begin{figure*}[t]
\centering
\includegraphics[width=0.99\linewidth]{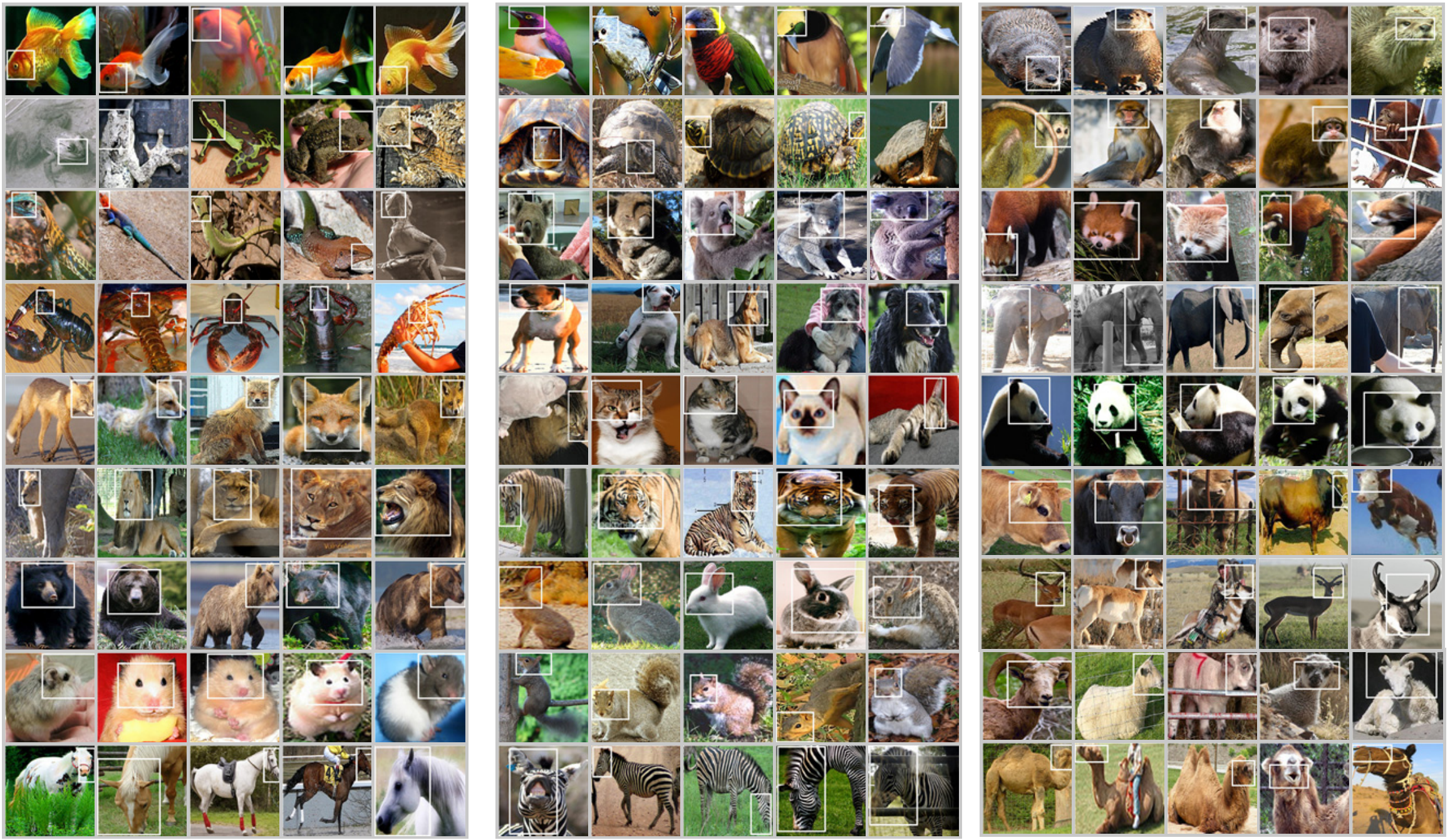}
\caption{Localization of the head part based on AOGs using the ILSVRC 2013 DET Animal-Part dataset~\cite{CNNAoG}. Users are given only \textbf{three} object images for interactive learning.}
\label{fig:results}
\end{figure*}

We also compared our method with two DPM-related methods, \emph{i.e.} the strongly supervised DPM (\textit{SS-DPM-Part})~\cite{SSDPM} and the technique in \cite{PLDPM} (\textit{PL-DPM-Part}), respectively. These two methods trained DPMs with part annotations for part localization. The next baseline, namely \textit{Part-Graph}, used a graphical model for part localization~\cite{SemanticPart}.

In the scope of weakly-supervised learning, ``simple'' methods are usually insensitive to the over-fitting problem. Therefore, we selected six baselines, which extracted \textit{fc7} features for each image patch and used a SVM for part detection. The first three baselines were original proposed in \cite{CNNAoG}, namely \textit{fc7+linearSVM}, \textit{fc7+RBF-SVM}, and \textit{fc7+NN}, which used a linear SVM, a RBF-SVM, and the nearest-neighbor strategy, respectively, for part detection. The other three baselines combined both the \textit{fc7} feature and the spatial position $(x,y)$ ($-1\leq x,y\leq1$) of each image patch as the final feature. We used \textit{fc7+sp+linearSVM}, \textit{fc7+sp+RBF-SVM}, and \textit{fc7+sp+NN} to denote the last three baselines.

All the baselines were provided with object bounding boxes and used the same set of part annotations for training to ensure a fair comparison.


\begin{table}[t]
\centering
\resizebox{1.0\linewidth}{!}{\begin{tabular}{l|cccc}
\hline
\!\!&\!\!\! bird beak \!\!&\!\!\! bird neck \!\!&\!\!\! bird wing \!\!&\!\!\! cat eye\\
{\footnotesize Mining-raw~\cite{CNNAoG}} \!\!&\!\!\! 0.1357 \!\!&\!\!\! 0.1822 \!\!&\!\!\! 0.1622 \!\!&\!\!\! 0.1586\\
Ours \!\!&\!\!\! {\bf0.1225} \!\!&\!\!\! {\bf0.1570} \!\!&\!\!\! {\bf0.1580} \!\!&\!\!\! {\bf0.1331}\\
\hline
\!\!&\!\!\! cow ear \!\!&\!\!\! dog nose \!\!&\!\!\! dog eye \!\!&\!\!\! horse ear\\
{\footnotesize Mining-raw~\cite{CNNAoG}} \!\!&\!\!\! 0.1819 \!\!&\!\!\! 0.1902 \!\!&\!\!\! 0.1273 \!\!&\!\!\! 0.2123\\
Ours \!\!&\!\!\! {\bf0.1725} \!\!&\!\!\! {\bf0.1789} \!\!&\!\!\! {\bf0.1032} \!\!&\!\!\! {\bf0.1739}\\
\hline
\!\!&\!\!\! horse eye \!\!&\!\!\! sheep neck \!\!&\!\!\! {\small sheep muzzle} \!\!&\!\!\! sheep eye\\
{\footnotesize Mining-raw~\cite{CNNAoG}} \!\!&\!\!\! 0.2229 \!\!&\!\!\! 0.1447 \!\!&\!\!\! 0.2203 \!\!&\!\!\! 0.1624\\
Ours\!\!&\!\!\! {\bf0.2046} \!\!&\!\!\! {\bf0.1359} \!\!&\!\!\! {\bf0.1786} \!\!&\!\!\! {\bf0.1293}\\
\hline
\end{tabular}}
\vspace{2pt}
\caption{Normalized distance for localization of different object parts on the Pascal VOC Part dataset. Localization results for the head has been shown in Table~\ref{tab:voc}.}
\label{tab:vocParts}
\end{table}

\subsection{Evaluation metric}

\cite{SemanticPart} mentioned that it is necessary to remove factors of object detection from the evaluation of part localization. Therefore, as in \cite{CNNAoG}, original images were cropped using object bounding boxes for testing to make a fair comparison. All the baselines used the cropped images for part localization/detection. In addition, some baselines for part detection may predict more than one locations for a part. Like in \cite{CNNSemanticPart,SemanticPart}, we took the detected part with the highest confidence in each image as the localization result. We evaluated part-localization performance using the metric of \textit{normalized distance}, which has been widely used~\cite{CNNSemanticPart,CNNAoG}. Given an object, the normalized distance is the Euclidean distance between predicted part center and ground-truth part center, divided by the diagonal length of the object bounding box. 


\subsection{Experimental results and analysis}

In experiments, we learned AOGs for the head, neck, nose, muzzle, beak, eye, ear, and wing parts of the six animal categories in the Pascal VOC Part dataset. For the ILSVRC 2013 DET Animal-Part dataset and the CUB200-2011 dataset, we learned an AOG for the head part\textcolor{red}{\footnote[7]{It is the ``forehead'' part for birds in the CUB200-2011 dataset.}} of each category. Because the head is shared by all categories in the two datasets, we selected the head as the target part to enable a fair comparison.

For each of the 37 object categories in the above three benchmark datasets, the experimental setting was the same as in \cite{CNNAoG}. Each part concept of the category was defined to have three different part templates, and a single part box was annotated for each part template. \emph{I.e.} we used a total of three annotations to build the AOG for the part. All the baselines learned models using the same three part annotations. On average, labeling a bounding box cost 3.4 seconds (\emph{i.e.} 6.8 seconds for labeling both the object and the part), and the average time of human interactions per image was 12.3 seconds.

Fig.~\ref{fig:holding} shows patterns before and after human interactions. Fig.~\ref{fig:results} shows part-localization results based on AOGs. Tables~\ref{tab:imgnet}, \ref{tab:cub200} and \ref{tab:voc} compare part-localization performance of different methods on the ILSVRC 2013 DET Animal-Part dataset, the CUB200-2011 dataset, and the Pascal VOC Part dataset, respectively. Table~\ref{tab:vocParts} lists localization results of various object parts in the Pascal VOC Part dataset before and after human-interactions. Our method exhibited superior performance to other baselines.

\section{Conclusions and discussion}

In this paper, we attempted to use human interactions to manually correct the representation of a semantic part, in the scenario of weak-supervised learning. We proposed to build a model for a semantic part by selecting related latent patterns from pre-trained CNNs and removing irrelevant patterns based on human subjective perception. We successfully mined and transferred latent patterns from a pre-trained CNN to a human-interpretable AOG model, which eased the visualization of latent patterns and allowed people to directly manipulate these patterns in the AOG. We can parallel this learning process to building LEGO blocks.

With the help of human interactions, our method ensured that the model used semantically correct patterns to represent an object part. This is different from conventional batch learning methods that usually let the computer ``guess'' the target knowledge representation from training samples. Our method exhibited superior performance in experiments, which demonstrated the effectiveness of incorporating human interactions in weakly-supervised learning.

{\small
\bibliographystyle{ieee}
\bibliography{TheBib}

\begin{thebibliography}{10}\itemsep=-1pt

\bibitem{CNNVisualization_5}
M.~Aubry and B.~C. Russell.
\newblock Understanding deep features with computer-generated imagery.
\newblock {\em In ICCV}, 2015.

\bibitem{SSDPM}
H.~Azizpour and I.~Laptev.
\newblock Object detection using strongly-supervised deformable part models.
\newblock {\em In ECCV}, 2012.

\bibitem{SemanticPart}
X.~Chen, R.~Mottaghi, X.~Liu, S.~Fidler, R.~Urtasun, and A.~Yuille.
\newblock Detect what you can: Detecting and representing objects using
  holistic models and body parts.
\newblock {\em In CVPR}, 2014.

\bibitem{ImageNet}
J.~Deng, W.~Dong, R.~Socher, L.-J. Li, K.~Li, and L.~Fei-Fei.
\newblock Imagenet: A large-scale hierarchical image database.
\newblock {\em In CVPR}, 2009.

\bibitem{FeaVisual}
A.~Dosovitskiy and T.~Brox.
\newblock Inverting visual representations with convolutional networks.
\newblock {\em In CVPR}, 2016.

\bibitem{FastRCNN}
R.~Girshick.
\newblock Fast r-cnn.
\newblock {\em In ICCV}, 2015.

\bibitem{CNNImageNet}
A.~Krizhevsky, I.~Sutskever, and G.~Hinton.
\newblock Imagenet classification with deep convolutional neural networks.
\newblock {\em In NIPS}, 2012.

\bibitem{CNN}
Y.~LeCun, L.~Bottou, Y.~Bengio, and P.~Haffner.
\newblock Gradient-based learning applied to document recognition.
\newblock {\em In Proceedings of the IEEE}, 1998.

\bibitem{PLDPM}
B.~Li, W.~Hu, T.~Wu, and S.-C. Zhu.
\newblock Modeling occlusion by discriminative and-or structures.
\newblock {\em In ICCV}, 2013.

\bibitem{CNNVisualization_2}
A.~Mahendran and A.~Vedaldi.
\newblock Understanding deep image representations by inverting them.
\newblock {\em In CVPR}, 2015.

\bibitem{MiningAOG}
Z.~Si and S.-C. Zhu.
\newblock Learning and-or templates for object recognition and detection.
\newblock {\em In PAMI}, 2013.

\bibitem{ObjectDiscoveryCNN_2}
M.~Simon and E.~Rodner.
\newblock Neural activation constellations: Unsupervised part model discovery
  with convolutional networks.
\newblock {\em In ICCV}, 2015.

\bibitem{CNNSemanticPart}
M.~Simon, E.~Rodner, and J.~Denzler.
\newblock Part detector discovery in deep convolutional neural networks.
\newblock {\em In ACCV}, 2014.

\bibitem{CNNVisualization_3}
K.~Simonyan, A.~Vedaldi, and A.~Zisserman.
\newblock Deep inside convolutional networks: Visualising image classification
  models and saliency maps.
\newblock {\em In arXiv:1312.6034v2}, 2013.

\bibitem{VGG}
K.~Simonyan and A.~Zisserman.
\newblock Very deep convolutional networks for large-scale image recognition.
\newblock {\em In ICLR}, 2015.

\bibitem{CUB200}
C.~Wah, S.~Branson, P.~Welinder, P.~Perona, and S.~Belongie.
\newblock The caltech-ucsd birds-200-2011 dataset.
\newblock Technical Report CNS-TR-2011-001, In California Institute of
  Technology, 2011.

\bibitem{CNNVisualization_1}
M.~D. Zeiler and R.~Fergus.
\newblock Visualizing and understanding convolutional networks.
\newblock {\em In ECCV}, 2014.

\bibitem{explanatoryGraph_arXiv}
Q.~Zhang, R.~Cao, Y.~Wu, and S.-C. Zhu.
\newblock Interpreting cnn knowledge via an explanatory graph.
\newblock {\em In arXiv:1708.01785}, 2017.

\bibitem{CNNAoG}
Q.~Zhang, R.~Cao, Y.~N. Wu, and S.-C. Zhu.
\newblock Growing interpretable part graphs on convnets via multi-shot
  learning.
\newblock {\em In AAAI}, 2017.

\bibitem{CNNSemanticDeep}
B.~Zhou, A.~Khosla, A.~Lapedriza, A.~Oliva, and A.~Torralba.
\newblock Object detectors emerge in deep scene cnns.
\newblock {\em In ICRL}, 2015.

\bibitem{CNNSemanticDeep2}
B.~Zhou, A.~Khosla, A.~Lapedriza, A.~Oliva, and A.~Torralba.
\newblock Learning deep features for discriminative localization.
\newblock {\em In CVPR}, 2016.

\bibitem{AllenAOG}
L.~Zhu, Y.~Chen, Y.~Lu, C.~Lin, and A.~Yuille.
\newblock Max-margin and/or graph learning for parsing the human body.
\newblock {\em In CVPR}, 2008.

\end{thebibliography}
}

\onecolumn
\section*{Appendix}

\begin{figure*}[h]
\centering
\includegraphics[width=0.9\linewidth]{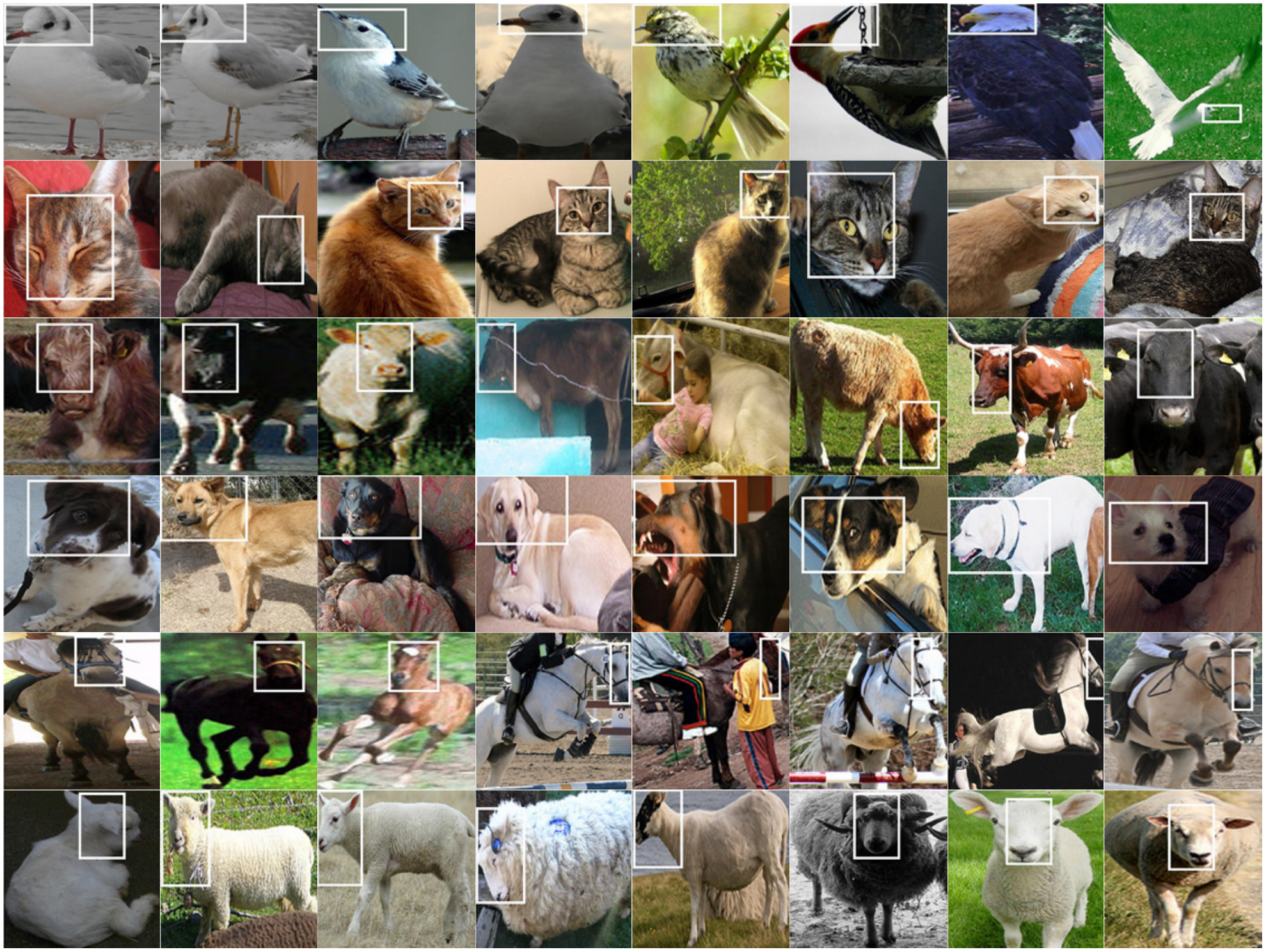}
\caption{Localization results of the head part on animal categories in the Pascal VOC Part dataset~\cite{SemanticPart}. Users are given only \textbf{three} object images for interactive learning.}
\end{figure*}

\begin{figure*}[t]
\centering
\includegraphics[width=0.9\linewidth]{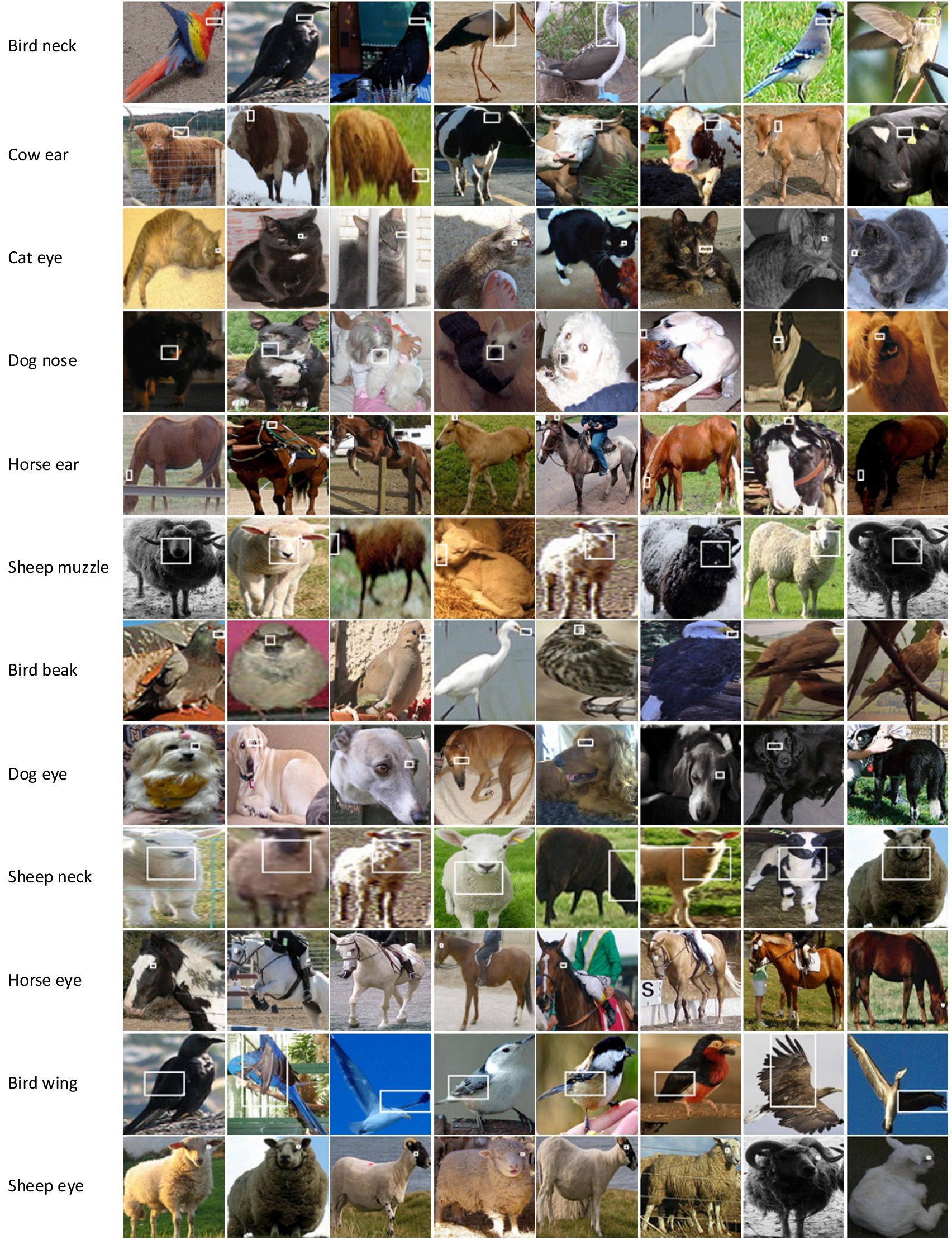}
\caption{Localization results of the neck, ear, eye, nose, muzzle, beak, and wing parts on animal categories in the Pascal VOC Part dataset~\cite{SemanticPart}. Users are given only \textbf{three} object images for interactive learning.}
\end{figure*}

\begin{figure*}[t]
\centering
\includegraphics[width=0.86\linewidth]{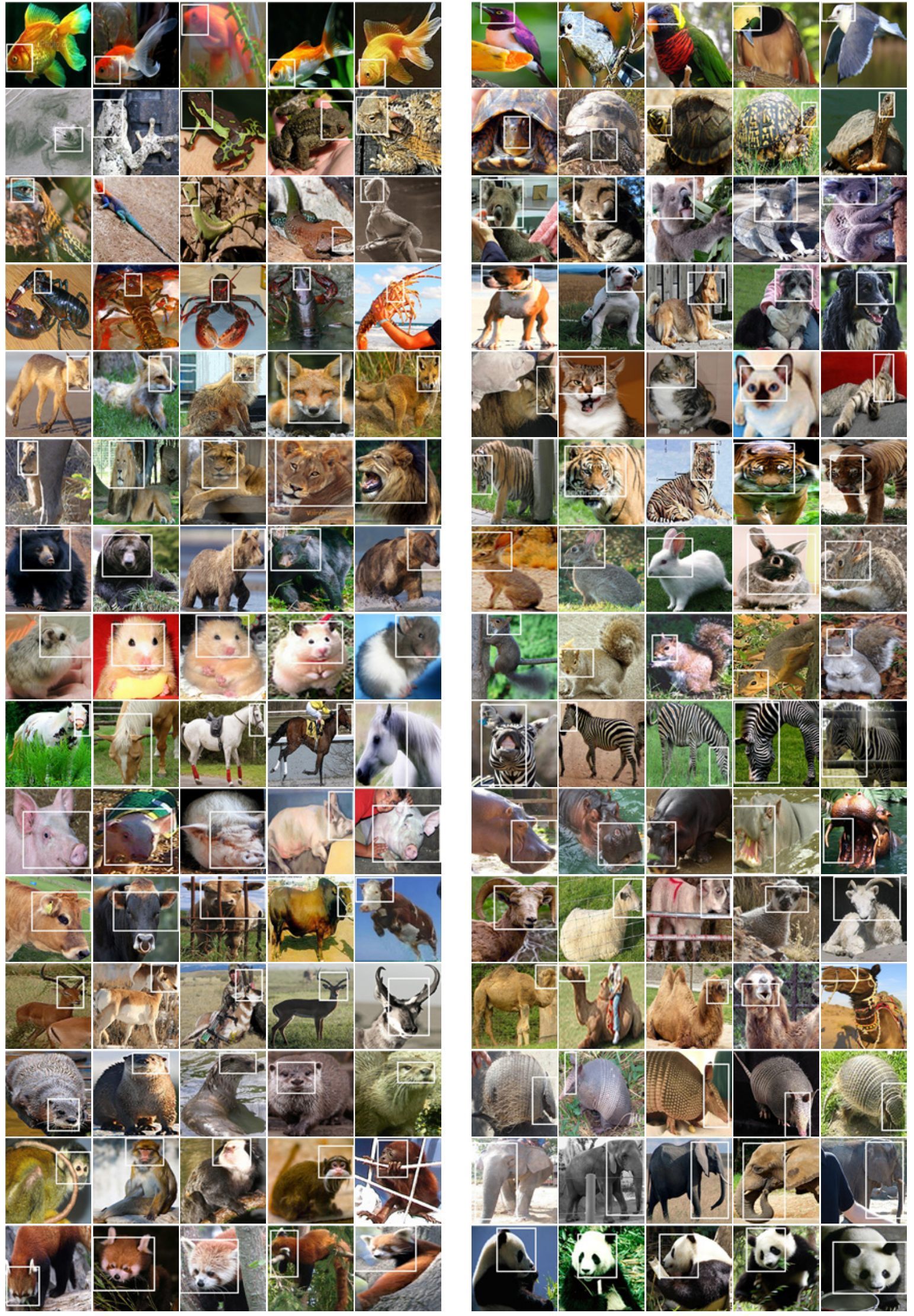}
\caption{Localization of the head part on the ILSVRC 2013 DET Animal-Part dataset~\cite{CNNAoG}. Users are given only \textbf{three} object images for interactive learning.}
\end{figure*}

\begin{figure*}[t]
\centering
\includegraphics[width=0.82\linewidth]{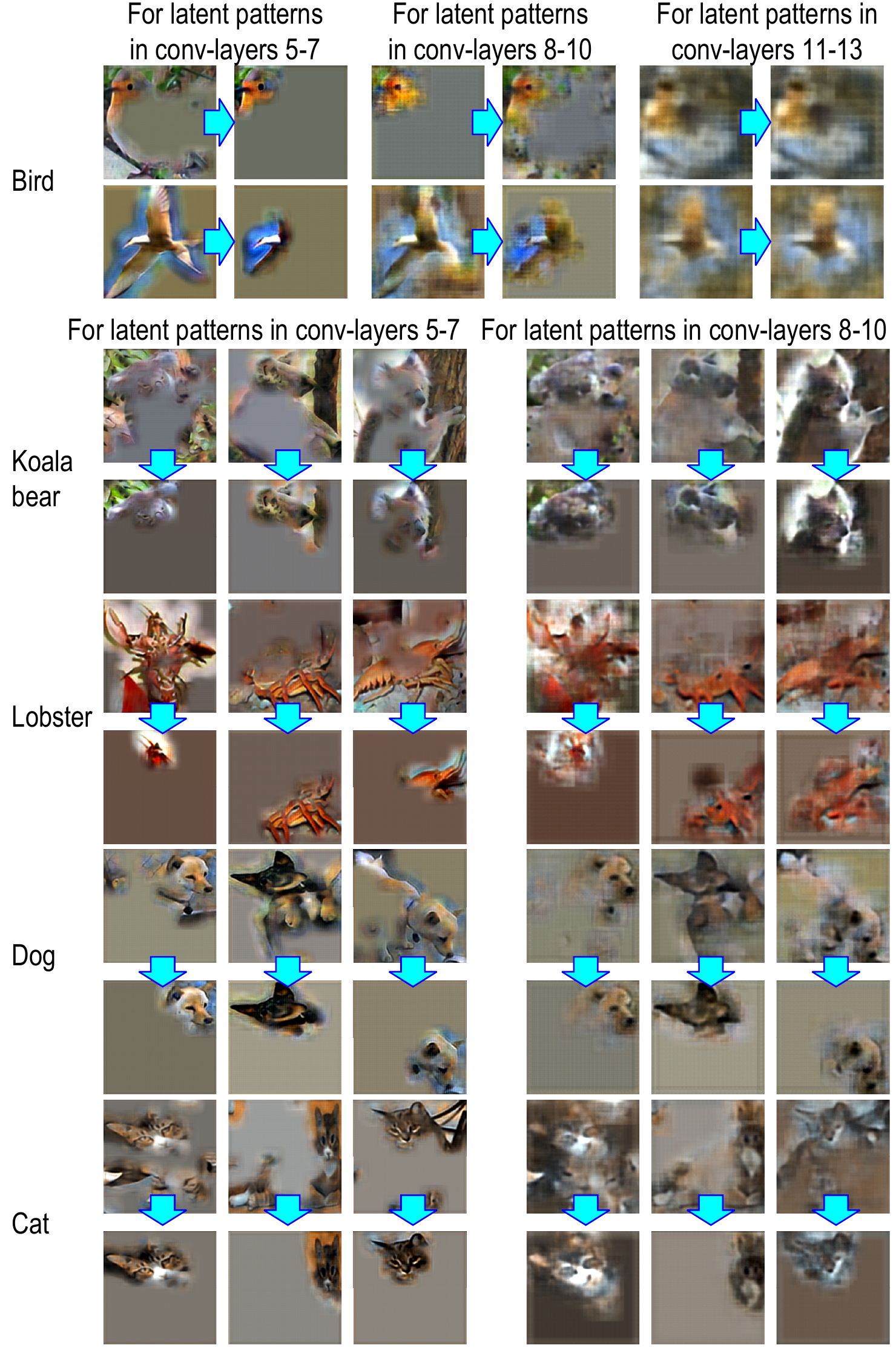}
\caption{Visualization of patterns for the head part before and after human interactions.}
\end{figure*}

\begin{figure*}[t]
\centering
\includegraphics[width=0.9\linewidth]{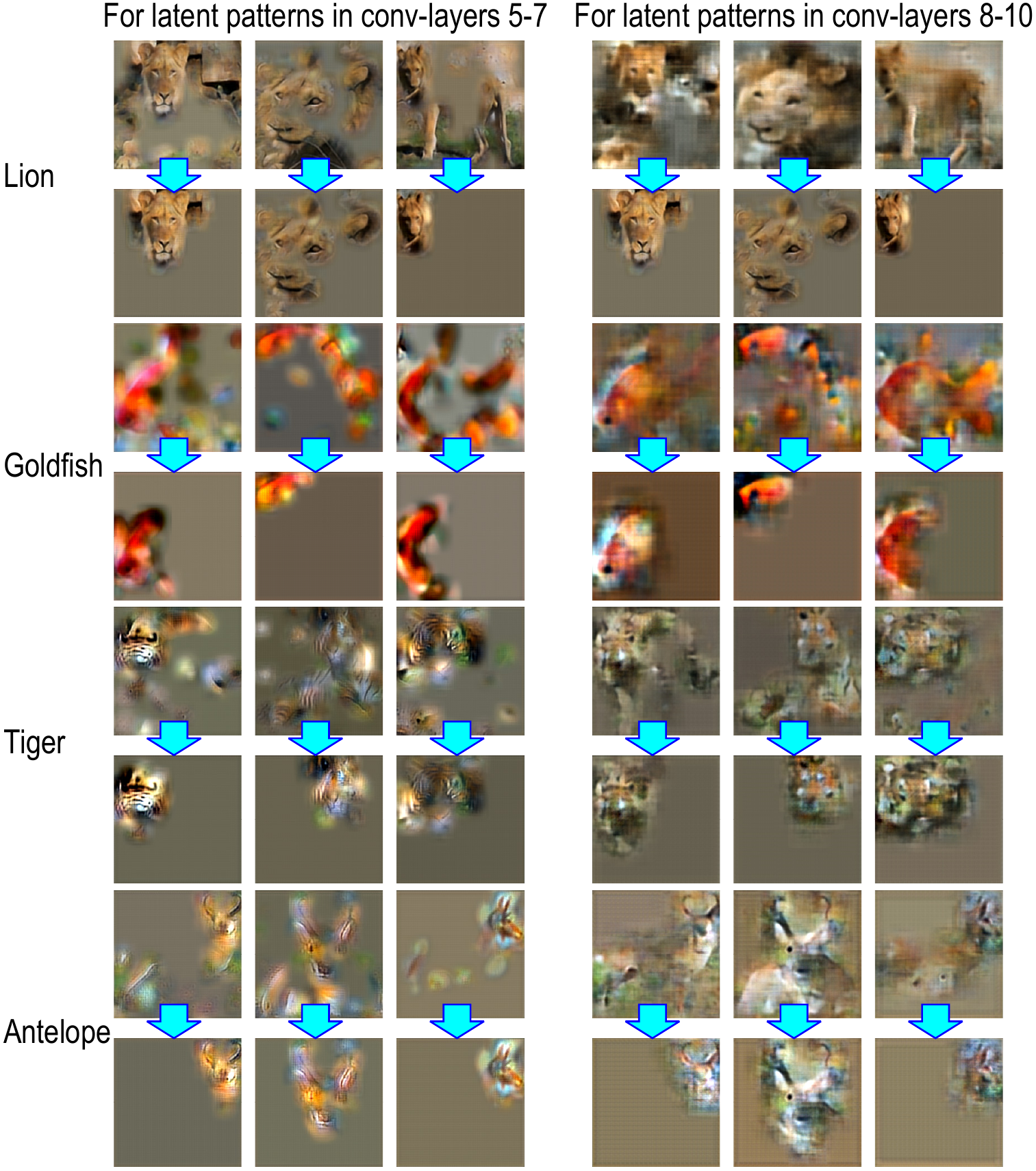}
\caption{Visualization of patterns for the head part before and after human interactions.}
\end{figure*}

\end{document}